\documentclass[10pt,twocolumn,letterpaper]{article}

\usepackage[pagenumbers]{cvpr} 

\usepackage[dvipsnames]{xcolor}
\definecolor{cvprblue}{rgb}{0.21,0.49,0.74}
\usepackage[pagebackref,breaklinks,colorlinks,citecolor=cvprblue]{hyperref}
\usepackage{tcolorbox}
\usepackage{multirow}

\newcommand{\myparagraph}[1]{
\textbf{#1} ---
}

\newcommand{\known}{\tcbox[on line,colframe=black,boxsep=0pt,left=1pt,right=1pt,top=1pt,bottom=1pt,boxrule=0.4pt]{Known task}}

\newcommand{\unknown}{\tcbox[on line,colframe=black,boxsep=0pt,left=1pt,right=1pt,top=1pt,bottom=1pt,boxrule=0.4pt]{Few-shot}}

\def\paperID{****} 
\def\confName{CVPR}
\def\confYear{2024}

\title{Task-conditioned adaptation of visual features in multi-task policy learning}

\author{
  Pierre Marza $^1$ \qquad
  Laetitia Matignon$^2$ \qquad
  Olivier Simonin$^1$ \qquad
  Christian Wolf$^3$ \qquad
  \\
  {\normalsize$^1$INSA Lyon} \quad
  {\normalsize$^2$UCBL} \quad
  {\normalsize$^3$Naver Labs Europe}  \quad
  \\
  {\tt \small \{pierre.marza, olivier.simonin\}@insa-lyon.fr} \\
  {\tt \small laetitia.matignon@univ-lyon1.fr}, {\tt \small christian.wolf@naverlabs.com} \\
  {\normalsize Project Page: \href{https://pierremarza.github.io/projects/task_conditioned_adaptation/}{https://pierremarza.github.io/projects/task\_conditioned\_adaptation/}}
}

\begin{document}
\maketitle

\begin{abstract}
\noindent
Successfully addressing a wide variety of tasks is a core ability of autonomous agents, requiring flexibly adapting the underlying decision-making strategies and, as we argue in this work, also adapting the perception modules. An analogical argument would be the human visual system, which uses top-down signals to focus attention determined by the current task. Similarly, we adapt pre-trained large vision models conditioned on specific downstream tasks in the context of multi-task policy learning. We introduce task-conditioned adapters that do not require finetuning any pre-trained weights, combined with a single policy trained with behavior cloning and capable of addressing multiple tasks. We condition the visual adapters on task embeddings, which can be selected at inference if the task is known, or alternatively inferred from a set of example demonstrations. To this end, we propose a new optimization-based estimator. We evaluate the method on a wide variety of tasks from the CortexBench benchmark and show that, compared to existing work, it can be addressed with a single policy. In particular, we demonstrate that adapting visual features is a key design choice and that the method generalizes to unseen tasks given a few demonstrations.
\end{abstract}

\begin{figure}[t]
    \centering
    \includegraphics[width=0.47\textwidth]{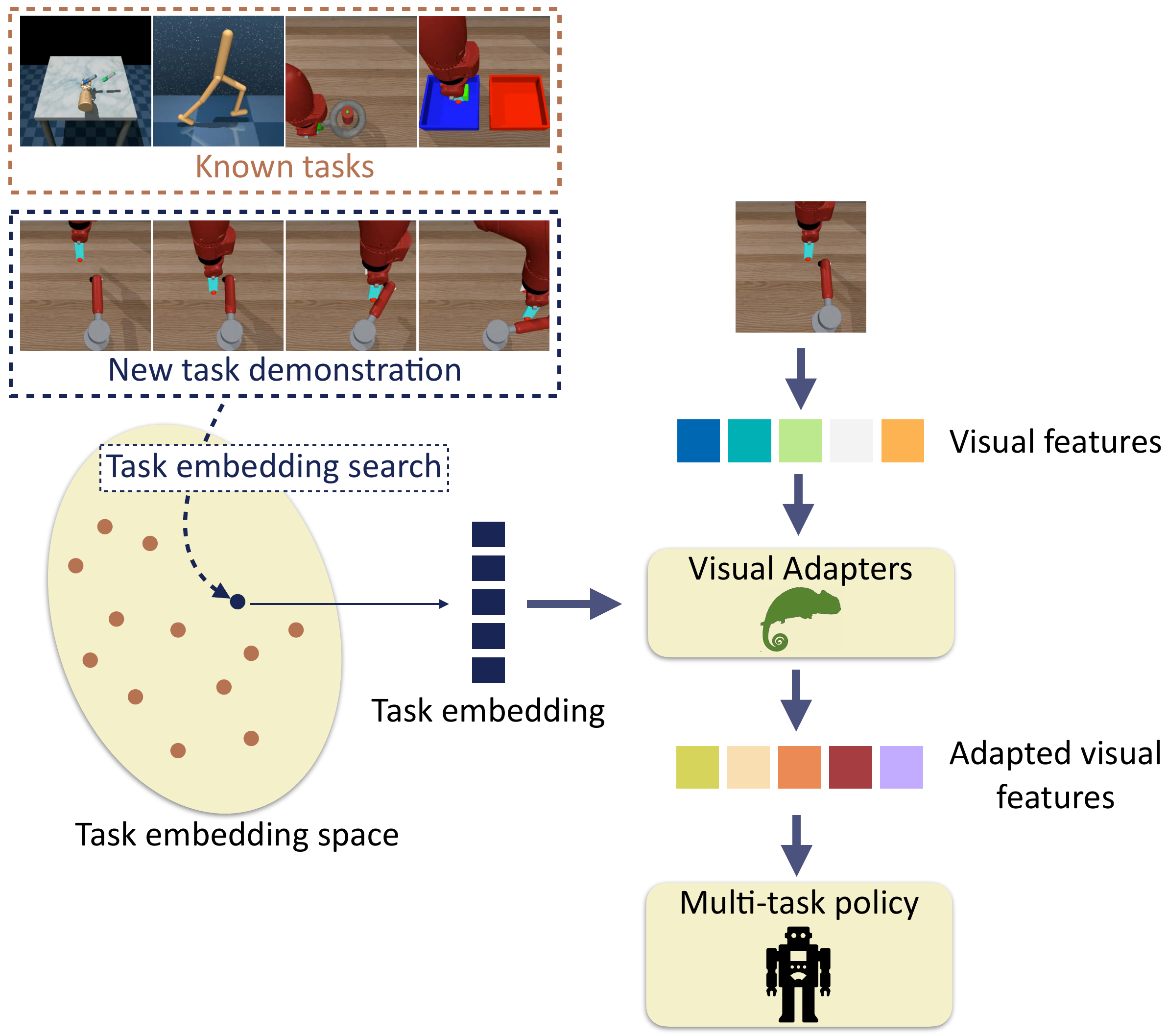}
    \caption{\label{fig:teaser-label} \textbf{Task-conditioned adaptation}: A single policy can be trained to address multiple heterogenous tasks including manipulation, legged motion etc., and few-shot learning is possible to address tasks given as demonstrations but unseen during training. A key element is the task-conditioned adaptation of visual features.}    
\end{figure}

\begin{figure*}[t]
    \centering
    \includegraphics[width=\textwidth]{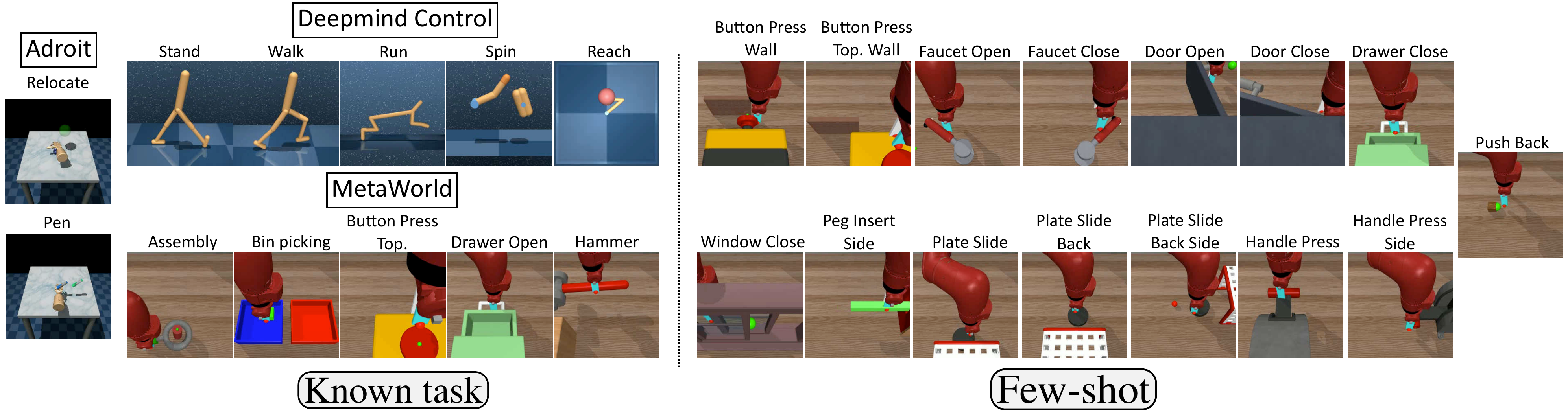}
    \caption{\label{fig:tasks}\textbf{Considered tasks}: We train the method on a set $T^k$ of known tasks and evaluate it either on the same set, with the task known (\known~ setting), or in a \unknown~ setting, where a new unseen task from a set $T^u$ is inferred from a few demonstrations.}  
\end{figure*}

\section{Introduction}
Vision is one of the most important modalities for agents interacting with the world and is almost indispensable for dexterous manipulation or locomotion, as no other sensor can provide information as rich and versatile. The inherent flexibility of the sensor comes with a high price, the high dimensionality of the information, and the complexity of the processes necessary to extract useful information. Humans and other biological agents are capable of adapting their perception systems to the task at hand. There is indeed evidence for bottom-up and top-down processes in human vision, with the latter guiding attention to regions determined by the requirements of the task \cite{CorbettaGoalDirected2002,BuschmanTopDown2007}.

There is a growing need for a similar versatility in artificial systems, and general neural networks have been trained from large-scale data in different domains such as natural language processing (NLP), computer vision (CV), and, more recently, robotics. A single general vision model coupled with a neural policy would be an appealing choice if it could allow an easy generalization to new domains or tasks. The wide adoption of attention mechanisms in several domains has made it easier for trained models to adapt their behavior to the requirements of different tasks without changing parameters, and it has been shown that attention plays a crucial role in specialization on a specific instance in language models \cite{voita2019analyzing} and vision and language models \cite{ReasoningCVPR2021}. However, even powerful generally pre-trained models can benefit from parameter adaptations to specific tasks, either through fine-tuning \cite{LORAICLR2022} or by adding additional trained adapter layers to a frozen model \cite{chen2022adaptformer}.

In robotics, prior work on the generalization capabilities of agents has focused on 
large-scale end-to-end training~\cite{reed_generalist_2022} or, targeting vision specifically, on pre-trained visual models required to generalize to various \textit{different} policies \cite{majumdar2023we,ma2022vip,nair2023r3m}. In this work, we argue and will show that a single policy can be trained for a large number of different tasks including manipulation, locomotion, and that the adaptation of visual features is highly beneficial, beyond the inherent adaptation capabilities of attention-based models. In particular, different tasks require diverse types of invariances and symmetries. While in principle it should be possible to learn to disentangle a sufficiently wide set of factors of variation in a captured representation such that it optimally performs on a wide variety of tasks, we will show that this is not the case for the arguably dominant pre-training method, masked auto-encoding (MAE)~\cite{he2022masked}.

We propose task-conditioned adaptation, which allows leveraging the high-quality representations of generally pre-trained large vision models, while keeping the required flexibility to address a wide variety of tasks, and also new (unseen) tasks. We introduce a set of task-conditioned visual adapters that can be inserted inside a pre-trained visual Transformer~\cite{vaswani2017attention}-based backbone. The task is characterized by an embedding space, which is learned from supervision during training. We show that this embedding space captures regularities of tasks and demonstrate this with \textit{few-shot capabilities}: the single policy and (adapted) visual representations can address new unseen tasks, whose embedding is estimated from a few demonstrations (cf. Figure \ref{fig:teaser-label}).

The contributions of this work can be summarized as follows: (i) task-conditioned visual adapters to flexibly modulate visual features to a specific task; (ii) a single multi-task policy solving tasks with different embodiments and environments; (iii) a task embedding optimization procedure based on a few demonstrations of a new task (unseen at training time) to adapt the model in a few-shot manner without any weight fine-tuning; (iv) quantitative and qualitative results assessing the gain brought by the different novelties.

\begin{figure}[t]
    \centering
    \includegraphics[width=0.47\textwidth]{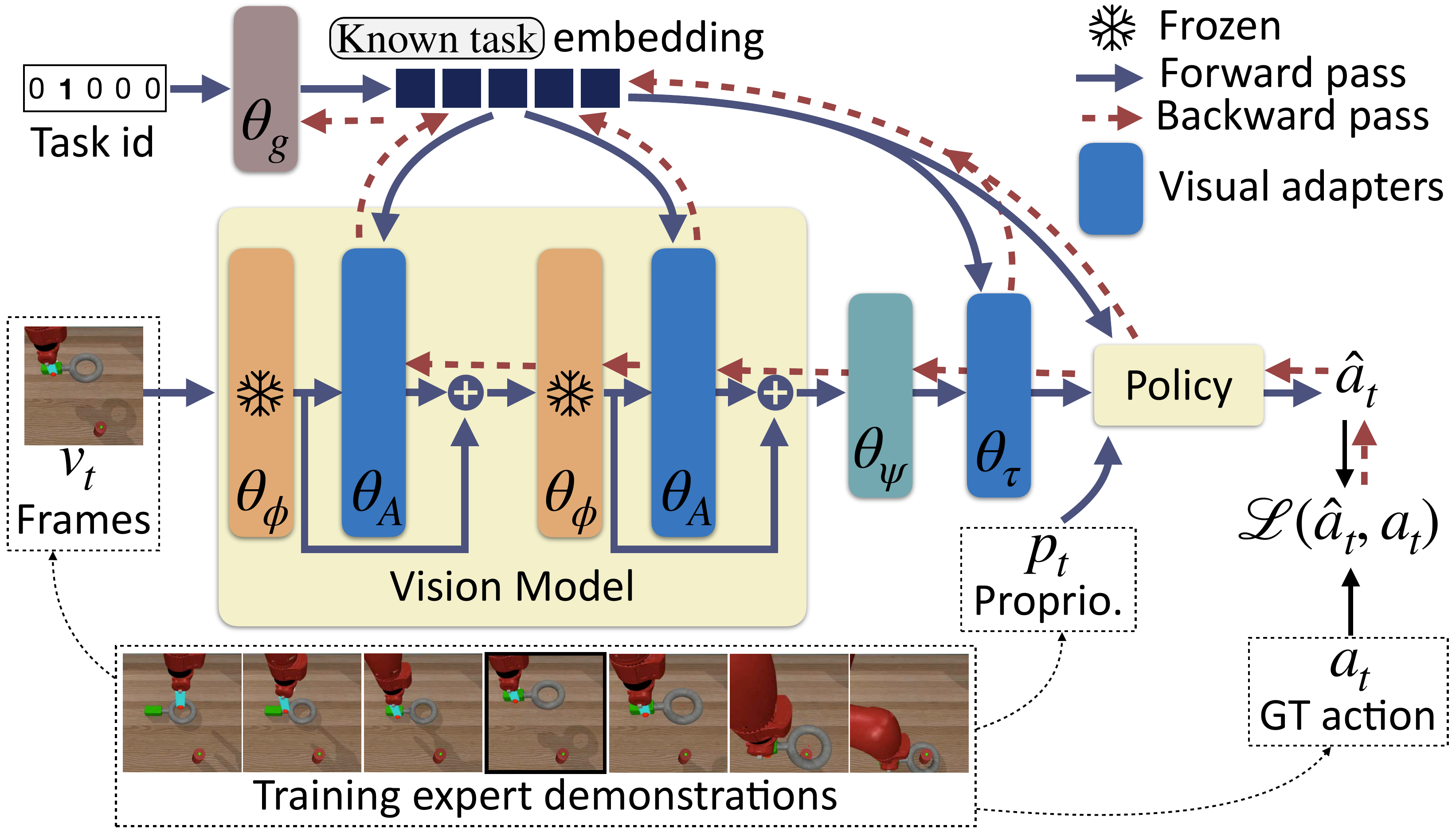} \\
    {\small (a) Training the policy and adapters in the \known~ setting.} \\
    \vspace{2ex}
    \includegraphics[width=0.47\textwidth]{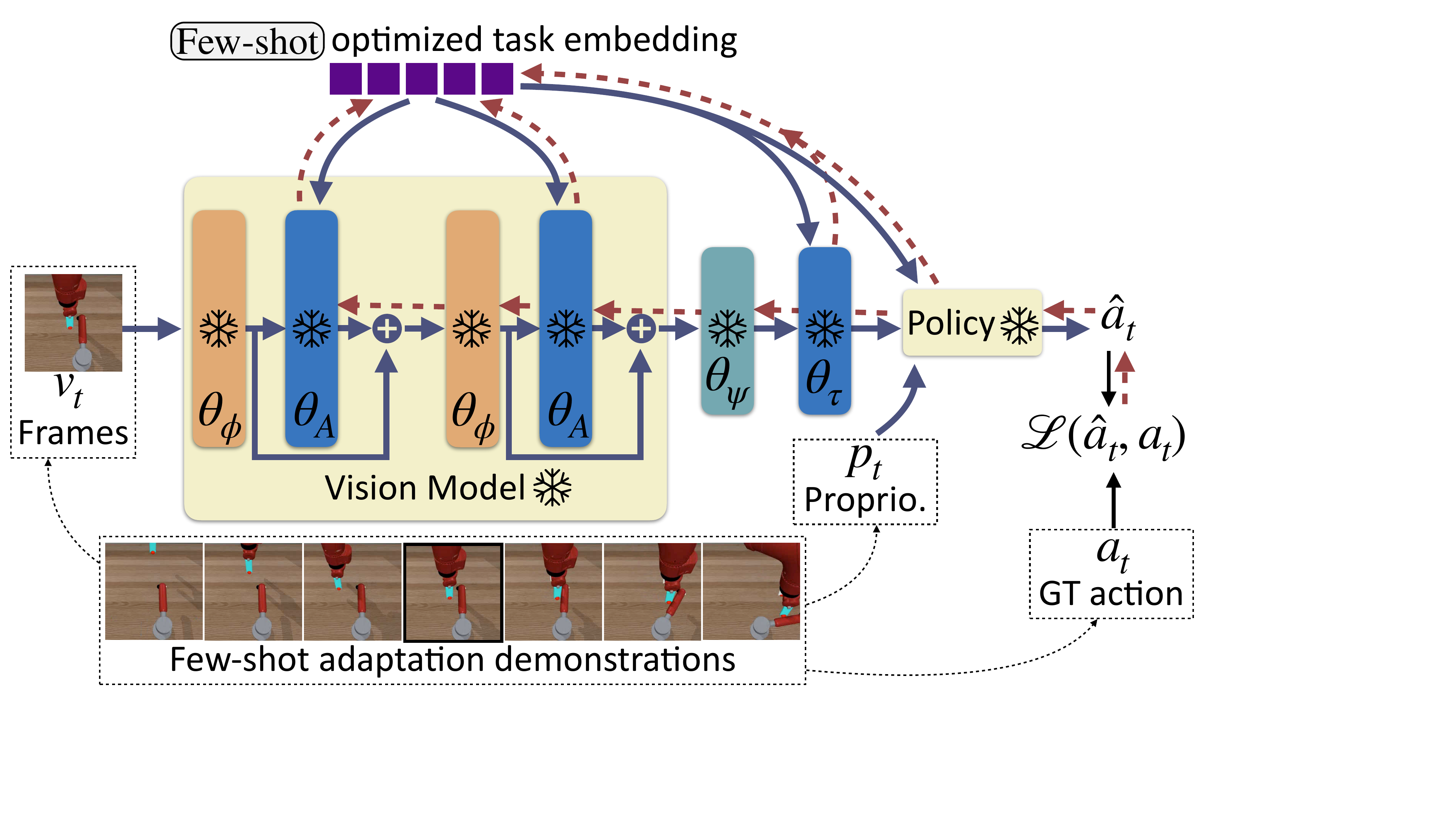} \\
    {\small (b) Optimization of the task embedding in the \unknown~ setting.} \\
    \vspace{2ex}
    \includegraphics[width=0.47\textwidth]{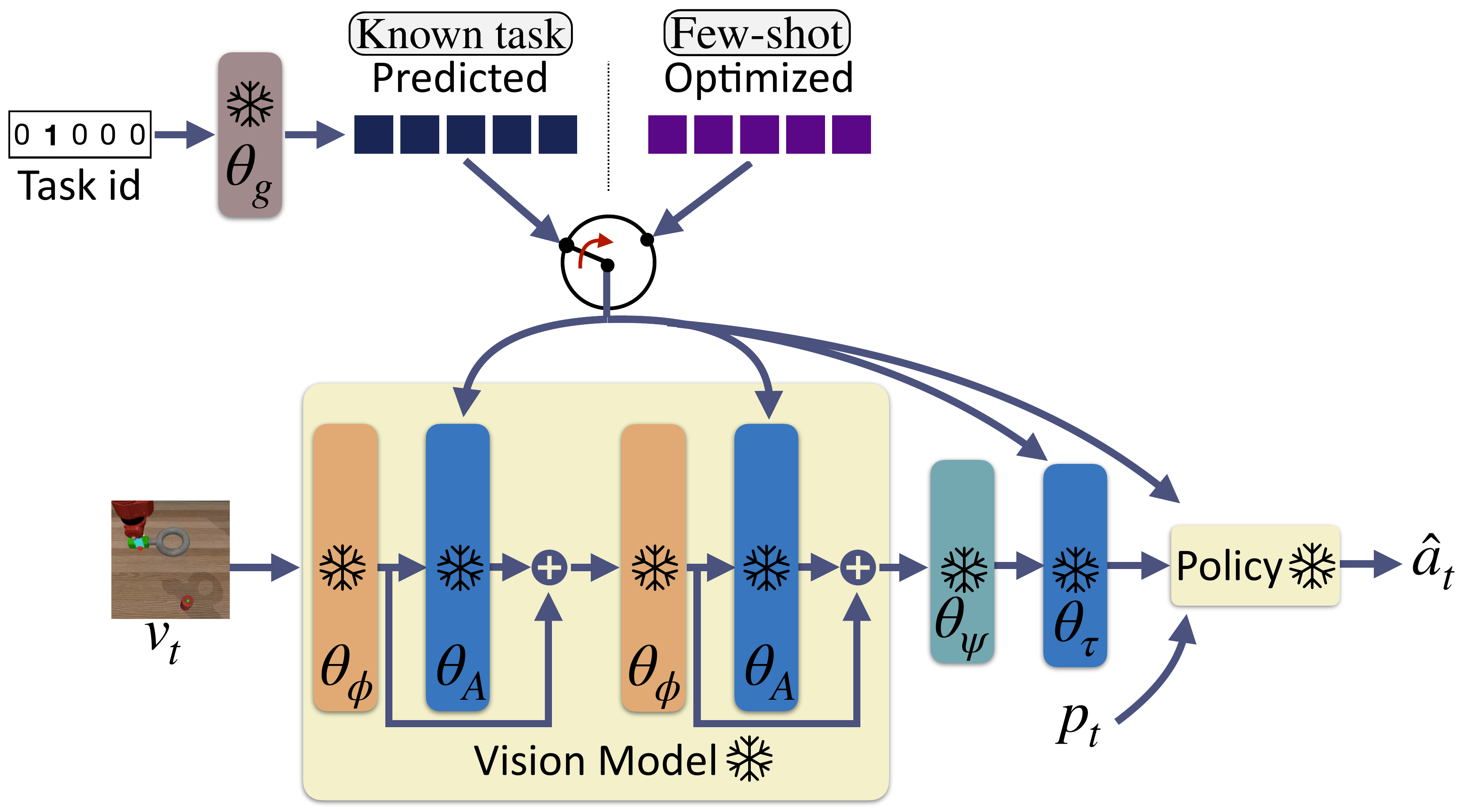} \\
    {\small (c) Inference for the settings: \known~ and \unknown.} \\
    \caption{\label{fig:method_figure} \textbf{Method overview}: (a) the adapted policy is trained with behavior cloning from expert demonstrations and given a visual encoder pre-trained with MAE. The model is conditioned on a task embedding learned from ground-truth 1-in-K task identifiers.  (b) In the \unknown~ case, a task embedding is estimated by optimization, maximizing the likelihood of given demonstrations of an unknown task. (c) Inference uses a task embedding predicted in the \known~ case, or optimized in the \unknown~ case.
    }
\end{figure}

\begin{figure*}[t]
    \centering
    \includegraphics[width=\textwidth]{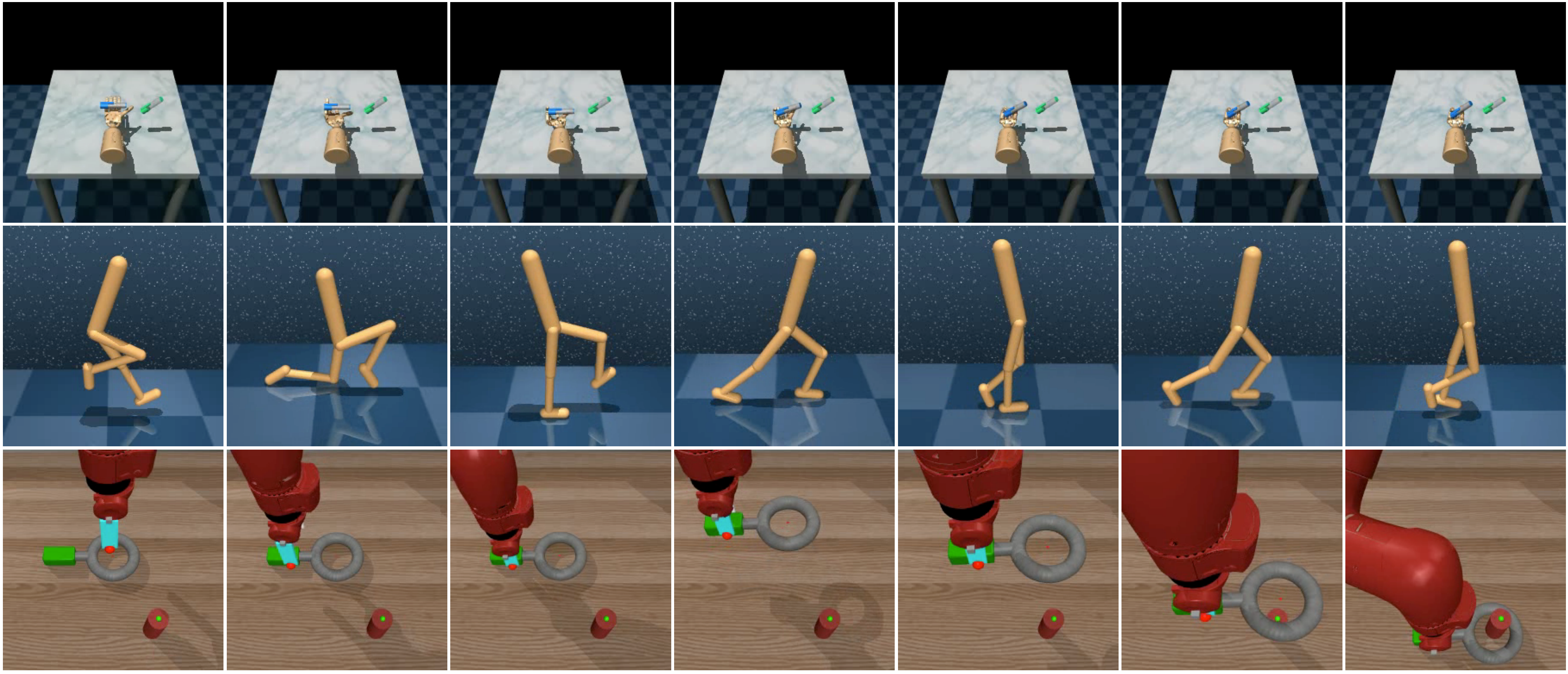}
    \caption{\label{fig:rollout_examples_known_tasks} \known~ --- \textbf{Qualitative results}: Three successful policy rollouts on known tasks from the test set. The multi-task approach performs well on a variety of diverse tasks while being trained on a limited set of demonstrations.}
\end{figure*}

\section{Related Work}
\myparagraph{Pre-trained visual representations for robotics} Backbone models pre-trained on large and diverse data have shown great promises in NLP~\cite{kenton2019bert, brown2020language, peters2018deep, liu2019roberta}, CV~\cite{radford2021learning, oquab2023dinov2, he2022masked, chen2020simple}, and more recently, robotics~\cite{majumdar2023we, parisi2022unsurprising, nair2023r3m, radosavovic2023real, ma2022vip, khandelwal2022simple, yadav2022offline, yadav2023ovrl}. \citet{parisi2022unsurprising} study visual pre-training methods for visuo-motor control, showing the quality of self-supervised representations. \citet{nair2023r3m} introduce R3M, a general vision model pre-trained on egocentric video data to capture temporal dynamics and semantic features, improving downstream manipulation performance. \citet{radosavovic2023real} employ the well-known MAE framework~\cite{he2022masked} to pre-train a single vision encoder applied to robots in the real world. \citet{ma2022vip} pre-train a visual model with a self-supervised value function objective on egocentric human videos to improve control policies. Finally, recent work~\cite{khandelwal2022simple, yadav2022offline, yadav2023ovrl} has shown the great promise of pre-trained models, either CLIP-based~\cite{khandelwal2022simple} or self-supervised~\cite{yadav2022offline, yadav2023ovrl}, in visual navigation.

The most related work is~\cite{majumdar2023we} which studies the impact of pre-trained vision models on a diversity of tasks gathered in a benchmark named \textit{Cortexbench}. They introduce a Vision Transformer (ViT)~\cite{dosovitskiy2020image} backbone, \textit{VC-1}, pre-trained from a set of out-of-domain datasets with a focus on egocentric visual frames. We believe that visual features should be task-dependent and study how the \textit{VC-1} model can be adapted to improve the performance of a single multi-task policy on a subset of \textit{Cortexbench} tasks, which is also different from previous work focusing on single-task policies.

\myparagraph{Transformer adapters} Transferring Transformer~\cite{vaswani2017attention}-based pre-trained models to new tasks or domains is an important topic. Methods involving adapter modules were introduced in NLP~\cite{houlsby2019parameter, pfeiffer2021adapterfusion, hu2021lora} to allow a fast and parameter-efficient transfer. Recent works employ the same methods in robotics~\cite{sharma2022lossless, liang2022transformer}. \citet{sharma2022lossless} insert visual adapters in a Vision Transformer (ViT)~\cite{dosovitskiy2020image} pre-trained model. They introduce different types of adapter blocks ("bottom", "middle", "top") located at diverse places in the visual model and show that combining them improves performance. \citet{liang2022transformer} also show the positive impact of inserting task-specific adapters, trained with imitation learning, in a pre-trained Transformer-based model to adapt to robotics tasks.

While prior work learns a specific set of adapters for each task, we argue that tasks share similarities and explore these regularities with a single set of task-conditioned adapters.

\myparagraph{Multi-task robotics policies} Having a single policy performing a wide range of tasks is a long-standing problem in robotics. With Deep Learning-based solutions becoming more popular, some prior work focuses on training multi-task neural agents. Approaches like BC-Z~\cite{jang2022bc}, RT-1~\cite{brohan2022rt},  RT-2~\cite{zitkovich2023rt} or Gato~\cite{reed2022generalist} study the scaling abilities of neural models to large-scale datasets. Trained generalist agents show strong performance on a wide set of tasks, and can generalize to some extent to novel tasks.

In contrast, our work leverages pre-trained vision models but does not assume access to a large set of robotics data. Instead, we focus on how to adapt visual features with reasonable computation requirements and train a single multi-task policy from only a few expert demonstrations. We also show promising few-shot adaptation to new unknown tasks without requiring very diverse training data. Somewhat related to our work is also TD-MPC2~\cite{hansen2023td}, which introduces a model-based RL algorithm to learn general world models and studies task embeddings to condition a multi-task policy. However, the latter does not act from vision while we specifically study how to modulate visual features conditioned by a task embedding.

\section{Task-conditioned adaptation}

All tasks considered in this work are sequential decision-making problems, where at each discrete timestep $t$ an agent receives the last $3$ visual frames as an observation $\mathbf{v}_t \in \mathbb{R}^{3 \times h \times w \times 3}$, where $h$ and $w$ are the height and width of images, and a proprioception input $\mathbf{p}_t \in \mathbb{R}^{d_a}$, and predicts a continuous action $\mathbf{\hat{a}}_t \in \mathbb{R}^{d_a}$, where $d_a$ is the dimension of the action space, which depends on the task at hand. 
We are provided with a training dataset of expert demonstrations to train a single policy, and for inference we study two different setups: \known, where we \textit{a priori} know the task to be executed, and \unknown, where the trained policy must be adapted to a new unseen task without fine-tuning only given a small set of demonstrations.

\myparagraph{Known tasks} Following~\cite{majumdar2023we}, we consider $K{=}12$ robotics tasks from $3$ benchmarks, Adroit~\cite{rajeswaran2018learning}, Deepmind control suite~\cite{tassa2018deepmind} and MetaWorld~\cite{yu2020meta}. The set of all known tasks is denoted as $T^k = \{t_i^k\}_{[i=1..K]}$, where $t_i^k$ is a 1-in-K vector encoding a known task, and is illustrated in Figure~\ref{fig:tasks}.

\myparagraph{Unknown tasks} The ability of our method to adapt to new skills is evaluated on a set of $U{=}15$ tasks from MetaWorld~\cite{yu2020meta}, for which we artificially generate demonstrations with a process described in section~\ref{Experiments}. The set of all unknown tasks is denoted as $T^u = \{t_i^u\}_{[i=1..U]}$, where $t_i^u$ is a 1-in-U vector encoding an unknown task, and is illustrated in Figure~\ref{fig:tasks}. Most importantly, $T^k \cap T^u = \emptyset$.

\subsection{Base agent architecture} 
Following a large body of work in end-to-end training for robotics, the agent directly maps pixels to actions and decomposes into a visual encoder and a policy.

\myparagraph{Visual encoder without adapters} following~\cite{majumdar2023we}, the visual encoder, denoted as $\phi$, is a ViT model~\cite{dosovitskiy2020image} pre-trained with masked auto-encoding (MAE). We keep pre-trained weights from VC-1 in~\cite{majumdar2023we}, which are publicly available. However, we change the way the representation is collected from the pre-trained model. Unlike~\cite{majumdar2023we}, the representation is not taken as the embedding of the 'CLS' token, which we consider to be undertrained by the MAE pretext task. Instead, we train a fully-connected layer $\psi$ to aggregate all the token representations of the last layer of the ViT except the 'CLS' token. The visual observation $\mathbf{v}_t$ associated with timestep $t$ is thus encoded as 
\begin{equation}
\mathbf{r}_t = \psi \bigr[ \phi(\mathbf{v}_t; {\theta}_{\phi}); {\theta}_{\psi} \bigr],
\end{equation}
where ${\theta}_{\phi}$ and ${\theta}_{\psi}$ are weights parametrizing $\phi$ and $\psi$ respectively. $\mathbf{r}_t \in \mathbb{R}^{3 \times d_r}$ as it contains the $d_r$-dim encoding of each of the $3$ last visual frames processed as a data batch, where $d_r$ is the output dimension of $\psi$.

As this will be relevant later, we recall here that a ViT $\phi$ is composed of a sequence of $N_l$ self-attention blocks, where $\phi_l$ is the layer at index $l$. If we denote the internal hidden representation predicted at layer $l$ as $\mathbf{s}^l_t$, and omit the weights of $\phi_l$ for simplicity, we have,

\begin{equation}
\mathbf{s}^l_t = \phi_l(\mathbf{s}^{l-1}_t),
\label{eq:vitlayer}
\end{equation}
where $\mathbf{s}^0_t{=}\mathbf{v}_t$.

\setlength{\tabcolsep}{1.5pt}
\begin{table*}
    \caption{\label{table:visual_adapters}\known~ --- \textbf{Impact of visual adapters}: Validation and test performance on known tasks of different baselines highlighting the gain brought by adapters. Both middle and top adapters bring a boost in performance, and conditioning them on the learned task embedding increases performance. Our multi-task policy outperforms single-task policies with VC-1 non-adapted features. \textbf{MT $\pi$}: multi-task policy -- NC: Non-conditioned -- C: Conditioned -- \textbf{Task emb.}: whether to input at evaluation time, either the learned task embedding and chosen from ground-truth (L), a random vector as the task embedding (Rd), or a randomly picked task embedding among the set of $K{=}12$ embeddings (RdP) -- Benchmarks avg: average performance across the $3$ considered benchmarks (Adroit, DMC, MetaWorld) -- Tasks avg: average performance across all $12$ known tasks. Performance is reported as \textit{mean $\pm$ std} over $3$ training runs (seeds).}
    \centering
    \small
    {
    \begin{tabular}{l c c c c c c c c c c c c c c c c}
        \toprule
        & \textbf{MT} & \multicolumn{2}{c}{\textbf{Adapters}} &  \textbf{Task} & \multicolumn{10}{c}{\textbf{Multi-task performance}} \\
         & \textbf{$\pi$}& Mid. & Top & \textbf{emb.} & \multicolumn{2}{c}{Adroit} & \multicolumn{2}{c}{DMC} & \multicolumn{2}{c}{MetaWorld} & \multicolumn{2}{c}{Benchmarks avg} & \multicolumn{2}{c}{Tasks avg} \\
         &&&&& Val & Test & Val & Test & Val & Test & Val & Test & Val & Test \\
         \midrule
         (a)~\cite{majumdar2023we} & $-$ & $-$ & $-$ & N/A & $44.0$ {\scriptsize $\pm$ 1.1} & $38.3$ {\scriptsize $\pm$ 2.5} & $49.6$ {\scriptsize $\pm$ 0.5} & $48.0$ {\scriptsize $\pm$ 0.3} & $53.5$ {\scriptsize $\pm$ 2.1} & $47.8$ {\scriptsize $\pm$ 2.0} & $49.1$ {\scriptsize $\pm$ 0.5} & $44.7$ {\scriptsize $\pm$ 0.3} & $50.3$ {\scriptsize $\pm$ 0.9} & $46.3$ {\scriptsize $\pm$ 0.4} \\
         \midrule
         (b) & $\checkmark$ & $-$ & $-$ & L & $36.3$ {\scriptsize $\pm$ 1.7} & $33.0$ {\scriptsize $\pm$ 4.0} & $55.3$ {\scriptsize $\pm$ 1.4} & $54.1$ {\scriptsize $\pm$ 0.3} & $41.7$ {\scriptsize $\pm$ 1.6} & $34.7$ {\scriptsize $\pm$ 0.9} & $44.4$ {\scriptsize $\pm$ 1.0} & $40.6$ {\scriptsize $\pm$ 1.8} & $46.5$ {\scriptsize $\pm$ 1.1} & $42.5$ {\scriptsize $\pm$ 1.2} \\
         \midrule
         (c) & $\checkmark$ & NC & $-$ & L & $40.2$ {\scriptsize $\pm$ 1.2} & $37.3$ {\scriptsize $\pm$ 2.8} & $54.0$ {\scriptsize $\pm$ 1.5} & $54.8$ {\scriptsize $\pm$ 1.9} & $45.8$ {\scriptsize $\pm$ 4.5} & $36.3$ {\scriptsize $\pm$ 2.5} & $46.7$ {\scriptsize $\pm$ 1.6} & $42.8$ {\scriptsize $\pm$ 1.9} & $48.3$ {\scriptsize $\pm$ 1.9} & $44.2$ {\scriptsize $\pm$ 1.8} \\ 
         (d) & $\checkmark$ & C & $-$ & L & $42.0$ {\scriptsize $\pm$ 2.5} & $\textbf{43.8}$ {\scriptsize $\pm$ 2.2} & $59.1$ {\scriptsize $\pm$ 1.3} & $58.8$ {\scriptsize $\pm$ 0.3} & $48.6$ {\scriptsize $\pm$ 4.8} & $40.8$ {\scriptsize $\pm$ 3.0} & $49.9$ {\scriptsize $\pm$ 2.1} & $47.8$ {\scriptsize $\pm$ 1.4} & $51.9$ {\scriptsize $\pm$ 2.1} & $48.8$ {\scriptsize $\pm$ 1.3} \\
         (e) & $\checkmark$ & C & NC & L & $\textbf{44.3}$ {\scriptsize $\pm$ 1.2} & $43.2$ {\scriptsize $\pm$ 1.5} & $\textbf{60.5}$ {\scriptsize $\pm$ 0.5} & $\textbf{60.3}$ {\scriptsize $\pm$ 2.5} & $58.6$ {\scriptsize $\pm$ 1.6} & $48.4$ {\scriptsize $\pm$ 1.9} & $54.5$ {\scriptsize $\pm$ 0.7} & $50.6$ {\scriptsize $\pm$ 0.8} & $57.0$ {\scriptsize $\pm$ 0.7} & $52.5$ {\scriptsize $\pm$ 1.1} \\
         (f) & $\checkmark$ & C & C & L & $42.0$ {\scriptsize $\pm$ 0.8} & $42.3$ {\scriptsize $\pm$ 1.0} & $59.9$ {\scriptsize $\pm$ 0.9} & $60.0$ {\scriptsize $\pm$ 0.5} & $\textbf{65.3}$ {\scriptsize $\pm$ 1.0} & $\textbf{54.5}$ {\scriptsize $\pm$ 3.3} & $\textbf{55.8}$ {\scriptsize $\pm$ 0.1} & $\textbf{52.3}$ {\scriptsize $\pm$ 1.0} & $\textbf{59.2}$ {\scriptsize $\pm$ 0.1} & $\textbf{54.8}$ {\scriptsize $\pm$ 1.2} \\
         \midrule
         (g) & $\checkmark$ & C & C & Rd & $4.2$ {\scriptsize $\pm$ 4.0} & $1.3$ {\scriptsize $\pm$ 0.9} & $10.3$ {\scriptsize $\pm$ 0.7} & $8.5$ {\scriptsize $\pm$ 1.1} & $1.3$ {\scriptsize $\pm$ 0.9} & $0.1$ {\scriptsize $\pm$ 0.1} & $5.3$ {\scriptsize $\pm$ 0.8} & $3.3$ {\scriptsize $\pm$ 0.2} & $5.5$ {\scriptsize $\pm$ 0.1} & $3.8$ {\scriptsize $\pm$ 0.4} \\
         (h) & $\checkmark$ & C & C & RdP & $0.7$ {\scriptsize $\pm$ 0.9} & $3.2$ {\scriptsize $\pm$ 2.5} & $5.7$ {\scriptsize $\pm$ 1.2} & $9.5$ {\scriptsize $\pm$ 6.1} & $0.9$ {\scriptsize $\pm$ 0.6} & $0.3$ {\scriptsize $\pm$ 0.4} & $2.4$ {\scriptsize $\pm$ 0.5} & $4.3$ {\scriptsize $\pm$ 2.2} & $2.9$ {\scriptsize $\pm$ 0.4} & $4.6$ {\scriptsize $\pm$ 2.5} \\
        \bottomrule
    \end{tabular}
    }
\end{table*}

\myparagraph{Single-task policy} following~\cite{majumdar2023we}, the policy $\pi$ is implemented as an MLP predicting actions from the input which is a concatenation of the current frame, two frame differences and the proprioception input $\mathbf{p_t}$,
\begin{equation}
\hat{\mathbf{a}}_t = \pi
\left (
\Bigr[ \mathbf{r}_{t,1}{-}\mathbf{r}_{t,0}, \mathbf{r}_{t,2}{-}\mathbf{r}_{t,1}, \mathbf{r}_{t,2}, \mathbf{p_t} 
\Bigr ] 
; {\theta}_{\pi}
\right ),
\label{eq:basepolicy}
\end{equation}
where $[ ~ ]$ is the concatenation operator and ${\theta}_{\pi}$ are weights parametrizing $\pi$.

\subsection{Adaptation}
Our key contributions are visual adapter modules along with a multi-task policy, which are all conditioned on the task at hand. This is done with a specific task embedding for each task, taken from an embedding space of dimension $d_e$, which is aimed to have sufficient regularities to enable few-show generalization to unseen tasks. Importantly, the different adapters and the multi-task policy are conditioned on the same task embedding, leading to a common and shared embedding space. For the \textit{known task} setting, where the ground-truth label of the task is available, the task embedding is projected from a 1-in-K vector with a linear function $g$ trained jointly with the adapters and the policy with the downstream loss (imitation learning). In the \textit{Few-shot} setting, at test time a new unknown task is described with a few demonstrations, and a task embedding is estimated through optimization, as will be detailed in subsection~\ref{te_optimization}. Figure \ref{fig:method_figure} outlines the architecture, its details will be given in the appendix.

Conditioned on a task embedding we denote as $\mathbf{e}$, the proposed adaptations are based on ``middle'' and ``top'' adapters following~\cite{houlsby2019parameter, sharma2022lossless}.

\myparagraph{Middle adapters} we add one trainable adapter after each ViT block to modulate its output. We introduce a set of middle adapters $A = \{\alpha_l\}_{[l=1..N_l]}$, where $\alpha_l$ is a 2-layer MLP. In the modified visual encoder $\phi^m$, each adapter modulates the output of the corresponding self-attention block and is conditioned on the task embedding $\mathbf{e}$. Its output is combined with the one of the self-attention layer through a residual connection. If we denote the internal hidden representation predicted at layer $l$ as $\mathbf{s}^{m, l}_t$,  and omit references to the weights of $\phi_l$ and $\alpha_l$ as in eq. (\ref{eq:vitlayer}) for simplicity, the associated forward pass of a given layer becomes,
\begin{align}
\mathbf{s}^{m, l}_t &= \phi^{m}_{l}(\mathbf{s}^{m, (l-1)}_t) \\
                    &= \phi_{l}(\mathbf{s}^{m, (l-1)}_t) +  \alpha_l(\phi_{l}(\mathbf{s}^{m, (l-1)}_t), \mathbf{e}).
\end{align}

\myparagraph{Top adapter} A top adapter $\tau$, also conditioned on the task at hand, is added after the ViT model, to transform the output of the aggregation layer $\psi$ to be fed to the multi-task policy (presented below). $\tau$ has the same architecture as a single middle adapter $\alpha_i$. The prediction of $\mathbf{r}_t^m$, equivalent to $\mathbf{r}_t$ in the non-adapted case, can be written as,
\begin{equation}
\mathbf{r}_t^m = \tau \Bigr[ \psi \bigr[ \phi^m(\mathbf{v}_t, \mathbf{e}; {\theta}_{\phi}, {\theta}_{A}); {\theta}_{\psi}  \bigr], \mathbf{e}; \theta_{\tau} \Bigr],
\end{equation}
where ${\theta}_{A}$ and ${\theta}_{\tau}$ are the weights parametrizing the middle adapters (each middle adapter has a different set of weights) and the top adapter respectively.

\myparagraph{Multi-task policy} We keep the architecture of the single-task policy in eq. (\ref{eq:basepolicy}), as in ~\cite{majumdar2023we}. However, instead of re-training a policy for each downstream task of interest, we train a single multi-task policy $ \pi^m$, whose action space is the union of the action spaces of the different tasks. During training we apply a masking procedure on the output, considering only the actions possible for the task at hand. 

Let's denote $\mathbf{\Tilde{r}}^m_t$ as the input to the policy derived from the adapted representation $\mathbf{r}^m_t$ and the proprioception input $\mathbf{p_t}$ as done in eq. (\ref{eq:basepolicy}). The conditioning on the task is done by concatenating $\mathbf{\Tilde{r}}^m_t$ with the task embedding  $\mathbf{e}$, giving

\begin{equation}
\hat{\mathbf{a}}_t = \pi^m
\left ( 
\left [ \mathbf{\Tilde{r}}^m_t, 
\mathbf{e}
\right ],
{\theta}_{\pi^m}
\right ),
\end{equation}
where ${\theta}_{\pi^m}$ are weights parametrizing $\pi^m$. 

\begin{figure*}[t]
    \centering
    \includegraphics[width=\textwidth]{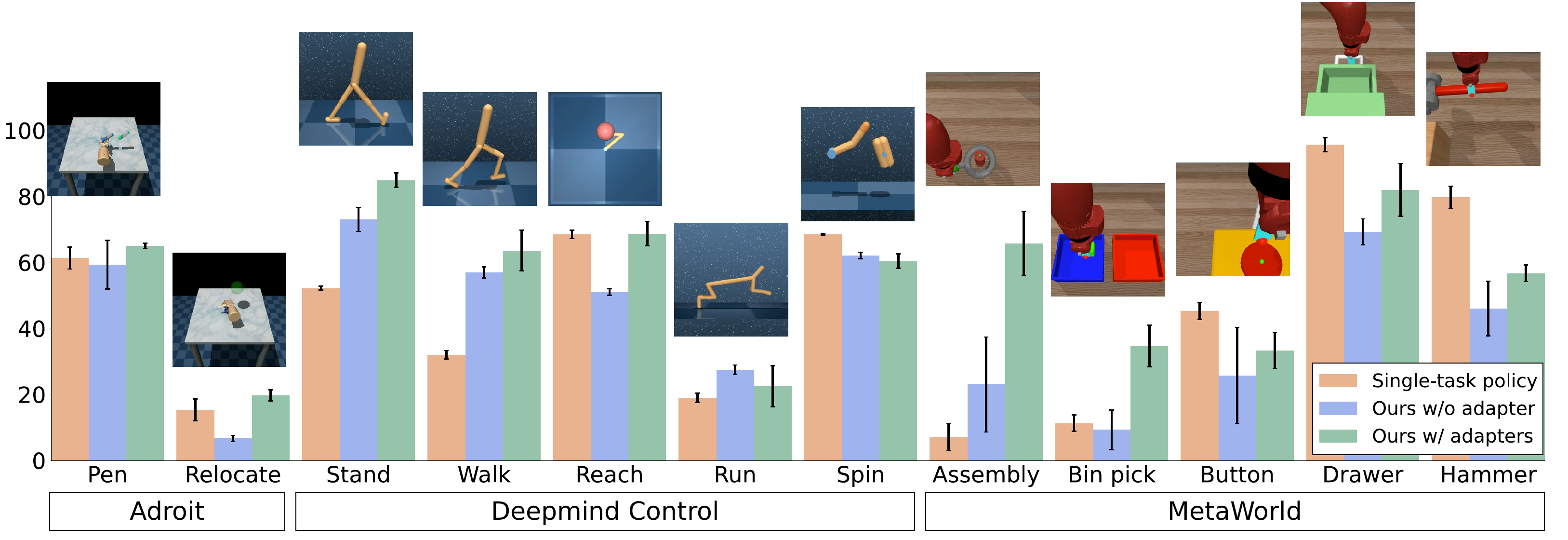}
    \caption{\label{fig:per_task_performance} \known~ --- \textbf{Per-task performance}  of policies in  Table~\ref{table:visual_adapters}: single-task policies (row (a)), our approach without any adapter (row (b)) and with conditioned middle and top adapters (row (f)). The adapters lead to a performance gain on most tasks, and our multi-task solution is competitive with single-task policies. Colored bars and error bars respectively show mean and std over $3$ training runs (seeds).}
\end{figure*}

\subsection{Training}
\label{paragraph-training} 
We train the model by keeping the weights of the pre-trained vision-encoder model $\theta_{\phi}$ frozen, only the weights of the adapter modules (${\theta}_{A}$, ${\theta}_{\tau}$), aggregation layer (${\theta}_{\psi}$), embedding layer ($\theta_g$) and multi-task policy (${\theta}_{\pi^m}$) are trained, cf. Figure~\ref{fig:method_figure}a. Lets' denote by $\Theta{=}{\{\theta}_{A}, {\theta}_{\tau}, {\theta}_{\psi}, {\theta}_{g}, {\theta}_{\pi^m}\}$ the set of optimized weights. We train with imitation learning, more specifically \textit{Behavior Cloning} (BC): for each known task $t_i^k$, we have access to a set of $N_i$ expert trajectories that are composed of $T_i$ discrete steps, including expert actions. The optimization problem is given as
\begin{equation}
\hat{\Theta} = \arg \min_{\Theta} \sum_{i=1}^{K} \sum_{n=1}^{N_i} \sum_{t=1}^{T_i}  \mathcal{L}(\hat{\mathbf{a}}_t^{i, n}, {\mathbf{a}}_t^{i, n}) ,
\end{equation} 
where $\hat{\mathbf{a}}_t^{i, n}$ and ${\mathbf{a}}_t^{i, n}$ are the predicted and ground-truth actions for a given step in a trajectory, and $\mathcal{L}$ is the Mean Squared Error loss.
\subsection{Few-shot adaption to new tasks}
\label{te_optimization} 
For the \unknown~ setting (cf. Figure~\ref{fig:method_figure}b), the task embedding $\mathbf{e}$ is unknown at inference and needs to be estimated from a set of $N_d$ example demonstrations $\mathcal{D}{=}\{ d_n \}_{[n=1..N_d]}$ where $d_n{=}\{ (\mathbf{v}^{n*}_t,\mathbf{p}^{n*}_t,\mathbf{a}^{n*}_t)\}_{[t=1..T_d]}$ is composed of observations and actions, with $T_d$ being the length of each demonstration. We exploit the conditioning property of the policy itself to estimate the embedding $\hat{\mathbf{e}}$ as the one which obtains the highest probability of the demonstration actions, when the policy is applied to the demonstration inputs, i.e. 
\begin{equation}
\hat{\mathbf{e}} = \arg \min_\mathbf{e}
\sum_{n=1}^{N_d}
\sum_{t=1}^{T_d}
\mathcal{L}(\pi^m
\left ( 
\left [ \mathbf{\Tilde{r}}^{mn*}_t, 
\mathbf{e}
\right ],
\theta_{\pi^m}
\right ), {\mathbf{a}}_t^{n*}),
\end{equation}
where $\mathbf{\Tilde{r}}^{mn*}$ is the representation extracted from the demonstration input $(\mathbf{v}^{n*}_t,\mathbf{p}^{n*}_t)$, and which itself depends on $\mathbf{e}$ (not made explicit in the notation).
The minimization is carried out with SGD from an embedding initialized to zero. It is important to note that only the task embedding is optimized, no neural weight is fine-tuned during the adaptation.

\begin{figure*}[t]
    \centering
    \includegraphics[width=\textwidth]{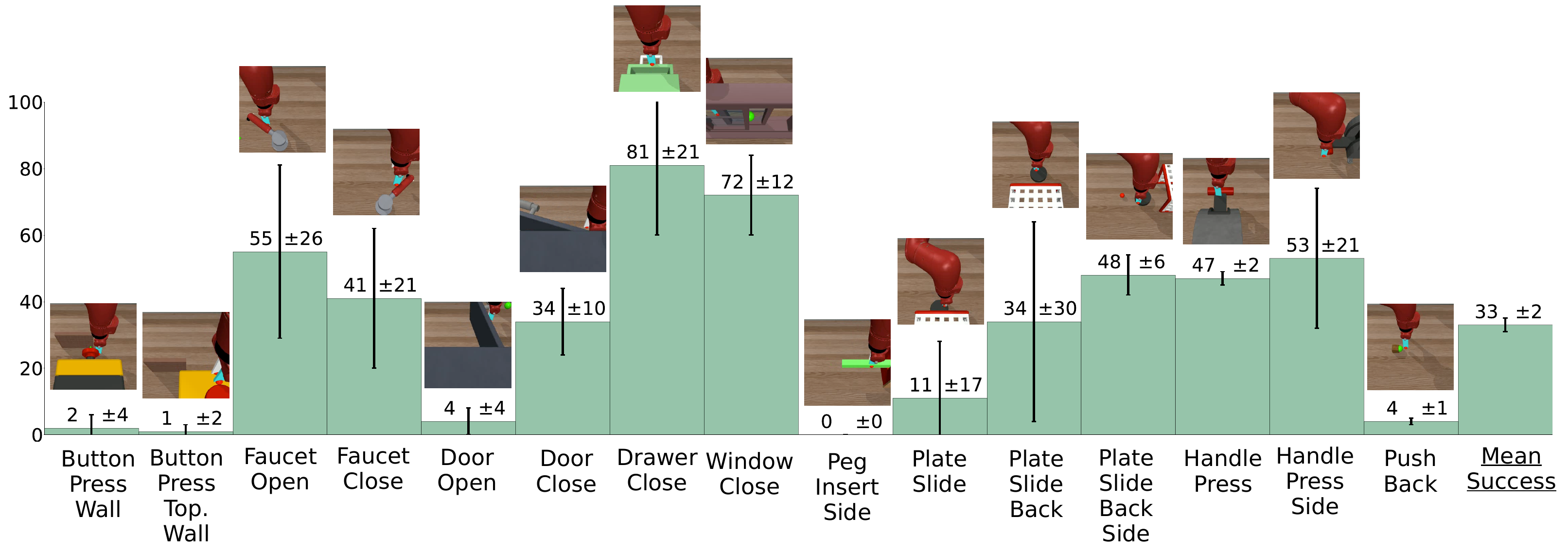}
    \caption{\label{fig:few_shot_per_task_performance} \unknown~ --- \textbf{Per-task performance}: After optimizing the task embedding for each task from 5 demonstrations, our method can adapt to many of them (without finetuning). Colored bars and error bars respectively show mean and std over $3$ training runs (seeds).}
\end{figure*}

\section{Experiments}
\label{Experiments}

\myparagraph{Training} all variants involving adapters and/or a multi-task policy (rows (b)-(f) in Table~\ref{table:visual_adapters}) were trained for $50$ epochs with behavior cloning, cf.~\S\ref{paragraph-training}, following training hyper-parameters in~\cite{majumdar2023we}. Between $20$ and $95$ expert trajectories are available depending on the task. We used the datasets of trajectories from~\cite{majumdar2023we}. 

\myparagraph{Evaluation} to better handle possible overfit on hyper-parameter selection, our evaluation setup is slightly different from~\cite{majumdar2023we} as we perform $100$ validation rollouts to select the best checkpoint of each model, and then test the chosen model on $100$ test rollouts. For our multi-task policy, the best checkpoint is the one with the highest average validation performance across all tasks. Single-task policies are validated only on the task they were trained on, giving them an advantage, and for this reason, they are reported as ``soft upper bounds''. We report the average performance and standard deviation among $3$ trained models (3 random seeds) for each variant as \textit{mean} $\pm$ \textit{std} (Table~\ref{table:visual_adapters}).

In total, we conduct evaluations of our method on $27$ different tasks, $12$ known and $15$ unknown, with varying environments, embodiments, and required sub-skills. This allows to evaluate the adaptation and generalization abilities of the multi-task policy.

\myparagraph{Evaluation metrics} 
Following~\cite{majumdar2023we}, we consider a rollout success ($1$ if the task was completed properly, $0$ otherwise) for tasks in the Adroit and MetaWorld benchmarks, and report the normalized return for DMC. Episodes have a maximum length of $1000$ steps and each step reward is comprised between $0$ and $100$ in DMC, normalization is therefore done by dividing the agent's return by $10$. For all tasks, performance is averaged across rollouts.

\myparagraph{\known~ --- Impact of visual adapters} Table~\ref{table:visual_adapters} presents a detailed comparison of different methods on the known task setting. 
The baseline in row (a) follows the setup in~\cite{majumdar2023we} to train single-task policies (one per task) from non-conditioned VC-1 features. For this variant, we use the representation of the 'CLS' token as the vector fed to the policy as done in~\cite{majumdar2023we}, while all other baselines use our proposed token aggregation layer.

Row (b) is our multi-task policy without any adapter. As expected there is a performance drop compared to the specialized policies in row (a), as the problem to solve has become more difficult. Adding adapters and conditioning them on the task embedding, shown in rows (c)-(f), brings a boost in performance, both for middle and top adapters. In particular, conditioning adds a further boost compared to non-conditioned adapters, with all choices enabled, row (f), obtaining the best average performance.

Rows (g) and (h) are ablation experiments evaluating the impact of choosing random task embeddings, row (g), or of taking a random choice between the 12 learned embeddings, row (h). In both cases, the performance collapses. 

A particularly important conclusion that can be drawn from the experiments outlined in Table~\ref{table:visual_adapters} is that the proposed multi-task approach (row (f)) outperforms the single-task policies without adapters (row (a)). This shows that a multi-task policy can perform well on a series of tasks while being trained on a limited set of demonstrations. Figure~\ref{fig:rollout_examples_known_tasks} presents $3$ successful test rollouts of our multi-task approach on diverse \textit{known tasks}.

Figure~\ref{fig:per_task_performance} visualizes the per-task test performance on \textit{known tasks} of single-task policies (row (a) in Table~\ref{table:visual_adapters}), our approach without any adapter (row (b) in Table~\ref{table:visual_adapters}) and with conditioned middle and top adapters (row (f) in Table~\ref{table:visual_adapters}). The proposed adapters lead to a performance gain on most tasks compared with the solution without adapters, and the multi-task solution is competitive with single-task policies, even outperforming them on half the tasks.

\begin{figure*}[t]
    \centering
    \includegraphics[width=0.9\textwidth]{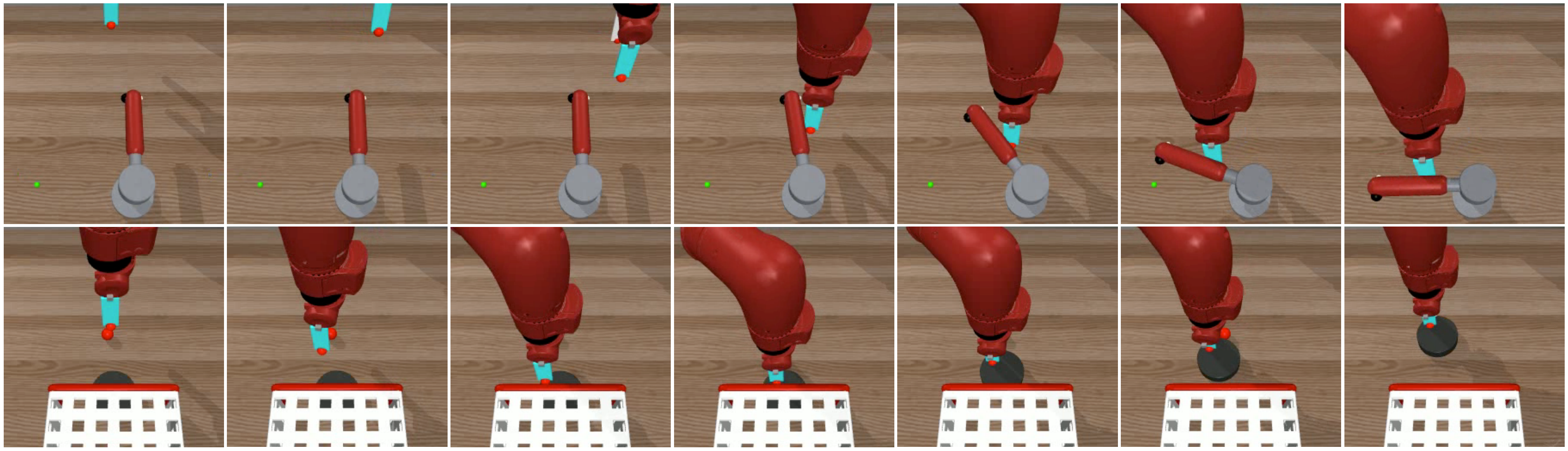} \\
    {(a) Successful rollouts} \\
    \vspace{1ex}
    \includegraphics[width=0.9\textwidth]{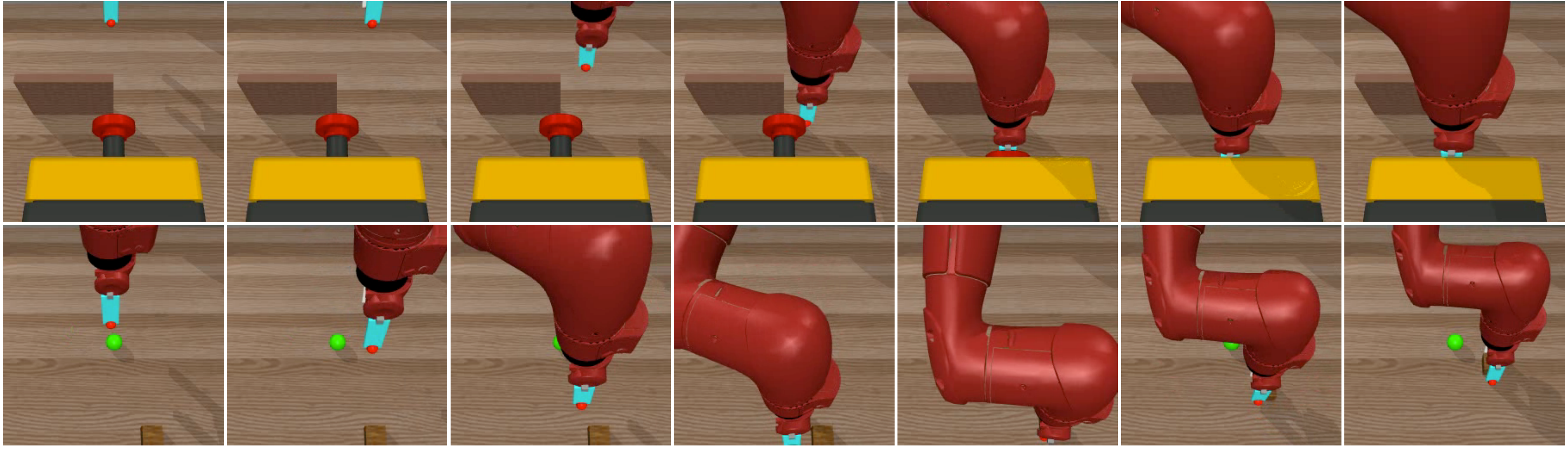} \\
    {(b) Failed rollouts} \\
    \caption{\label{fig:rollout_examples_unknown_tasks}\unknown~ --- \textbf{Qualitative results}: (a) The policy tackles new tasks involving objects and/or manipulation requirements unseen during training. (b) In the first row (\textit{button-press-wall} task), it performs the task correctly until the end where it fails to properly push the button fully. In the second row (\textit{push-back} task), it properly moves the cube but fails to bring it to the goal position (green dot).}
\end{figure*}

\begin{figure*}[t]
    \centering
    \includegraphics[width=\textwidth]{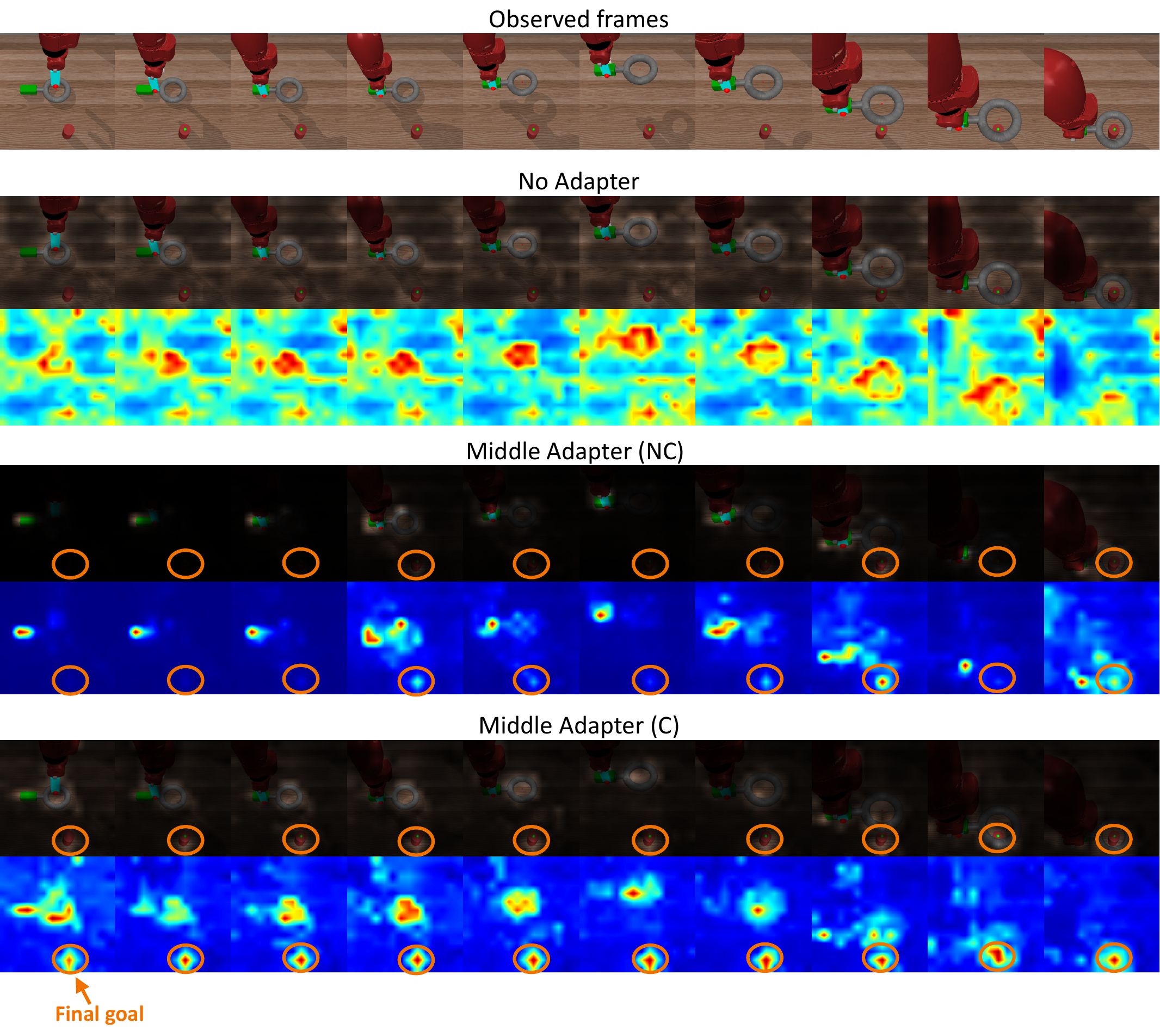}
    \caption{\label{fig:viz_attention_assembly} \textbf{Visualization of attention maps (Assembly task)}. First row: observed input frames. Following blocks: for each model type, we show the attention map of the last ViT layer, first overlaid on top of the visual frame and below as a colored heatmap. In this example, middle adapters allow to focus the attention on important regions, and task conditioning leads to a better covering of entire objects and agent parts, along with greater attention towards the final goal for all frames.}
\end{figure*}

\myparagraph{\known~ --- Additional ablation studies} are presented in the appendix (Sections B and C). Section~B shows that conditioning the policy when using task-conditioned adapters is not necessary, that our aggregation layer works better than using the 'CLS' token as input to the policy and further highlights the impact of both middle and top adapters. Section~C shows that our adapters improve visual embeddings extracted by two state-of-the-art pre-trained backbones, i.e. PVR~\cite{parisi2022unsurprising} and MVP~\cite{radosavovic2023real}, confirming the conclusions drawn from experiments on VC-1.

\myparagraph{\unknown~} adaptation to new tasks without finetuning model weights is an important ability of any general policy, which we evaluate with the following experiments: for each task within a set of unknown tasks (cf. Figure~\ref{fig:tasks}), we collect only $5$ demonstrations used to optimize a task embedding specific to this task with the method detailed in section \S\ref{te_optimization}. We then evaluate the method conditioned on the optimized embedding on $100$ test rollouts.

To generate the set $T^u$ of unknown tasks, we select tasks from the MetaWorld dataset that do not belong to CortexBench, and are thus not part of the set of training known tasks $T^k$. We collect demonstrations using single-task policies from TD-MPC2~\cite{hansen2023td} that were specifically trained on each task of MetaWorld independently. To ensure high-quality demonstrations, we only consider tasks where TD-MPC2 policies reach a success rate higher than $95\%$. Furthermore, to be compatible with the setup in CortexBench authors, in particular, to keep the same camera locations, we filter out the tasks where the goal is not always visible in the camera FOV. This leads to a set of $15$ unknown tasks that are quite different from the tasks in the training set $T^k$ as they involve different objects (\textit{handle press, faucet, plate, door, window}, etc.) and types of manipulation (\textit{sliding an object, lowering a press, opening a window}, etc.). Each collected demonstration is a sequence of visual frames, proprioception inputs, and expert actions. The optimization of the task embedding is performed independently for each task (cf. \S\ref{te_optimization}). We use the AdamW optimizer and a learning rate of $1e{-1}$ during the task embedding search.

Figure~\ref{fig:few_shot_per_task_performance} presents the per-task performance in this setting. Despite the large variations between the new tasks in $T^u$ and the ones in the training set ($T^k$), the multi-task policy can adapt to many of them, without requiring any weight finetuning. Interestingly, the method performs particularly well on the \textit{Drawer Close} task, which could be related to the presence of the time-inversed \textit{Drawer Open} task in the training set. This provides some evidence that the method can exploit regularities between tasks, which seem to be captured by the task embedding space, making it possible to generalize to unseen variations. Figure~\ref{fig:rollout_examples_unknown_tasks} (a) shows qualitative examples of successful rollouts on the new tasks. The policy manipulates new objects (\textit{faucet, plate, window}) and performs new moves (rotating the faucet or sliding the plate) not seen during training.

Finally, Figure~\ref{fig:rollout_examples_unknown_tasks} (b) shows failure cases on unknown tasks. As seen on the first row, the policy avoids the wall, reaches the button, and starts pushing it, but fails to push it fully. This particular behavior was also observed on other rollouts, explaining the low success rate on this \textit{button-press-wall} task while mastering a part of the required sub-skills. On the second row, the policy is able to move the cube but fails to bring it to the goal location (green dot). This gives some indication of the difficulty of the few-shot generalization case: exploiting regularities in the task space requires that tasks be performed more than just approximately, as often the success metric is sparse, and rollouts only count to the metric when they are executed fully and correctly.

\myparagraph{\known~ --- Visualizing the influence of task-conditioned adaptation} Sequences of visual frames used in this experiment are taken from a held-out set of expert trajectories not used at training time.  We visualize here the attention map of the last layer of the vision encoder. To this end, we sum attention maps for all tokens and all heads, and normalize them between $0$ and $1$. These visualizations are shown in Figures~\ref{fig:viz_attention_assembly}. The first row shows a sequence of visual frames and below, for each model variant (No Adapter, Middle Adapter (NC), Middle Adapter (C)), one can see the attention map overlaid on top of the visual frame and displayed below as a colored heatmap. As can be seen, the middle adapters help focus the attention on the most important parts of the image compared with vanilla VC-1 attention without adapters. When adapters are not conditioned (NC), they tend to produce very narrow attention. Conditioning on the task at hand keeps the focus on important regions and leads to covering the entire objects of interest and important agent parts. Most importantly, Figure~\ref{fig:viz_attention_assembly} shows that, when adapters are conditioned on the task embedding, more attention is put on the final goal in all frames, while this is not the case for unconditioned adapters. Another visualization is shown in Figure~\ref{fig:viz_attention_relocate} in the appendix.

\section{Conclusion}
Perception and action are closely tied together, and studies of human cognition have shown that \textit{a priori} knowledge about a downstream task guides the visual system. We follow this direction in the context of artificial agents by introducing task-conditioned adapters that modulate the visual features of a pre-trained neural backbone. Such adapters, conditioned on a learned task embedding, improve the performance of a multi-task policy across benchmarks and embodiments. Even more interesting is the use of task embeddings to adapt in a few-shot manner, i.e. from a small set of demonstrations, to new tasks unseen at training time. We propose an optimization procedure to estimate a new task embedding and achieve generalization to unseen tasks, involving new objects and manipulation sub-skills, providing evidence for regularities in the learned embedding space.

\textbf{Acknowledgement} ---
We thank ANR for support through AI-chair grant ``Remember'' (\small{ANR-20-CHIA-0018}).

\bibliographystyle{ieeenat_fullname}
\bibliography{main}

\clearpage
\appendix

\begin{center}
{\Large \textbf{Appendix}}
\end{center}

\section{Validation performance curves}
\begin{figure*}[t]
    \centering
    \includegraphics[width=\textwidth]{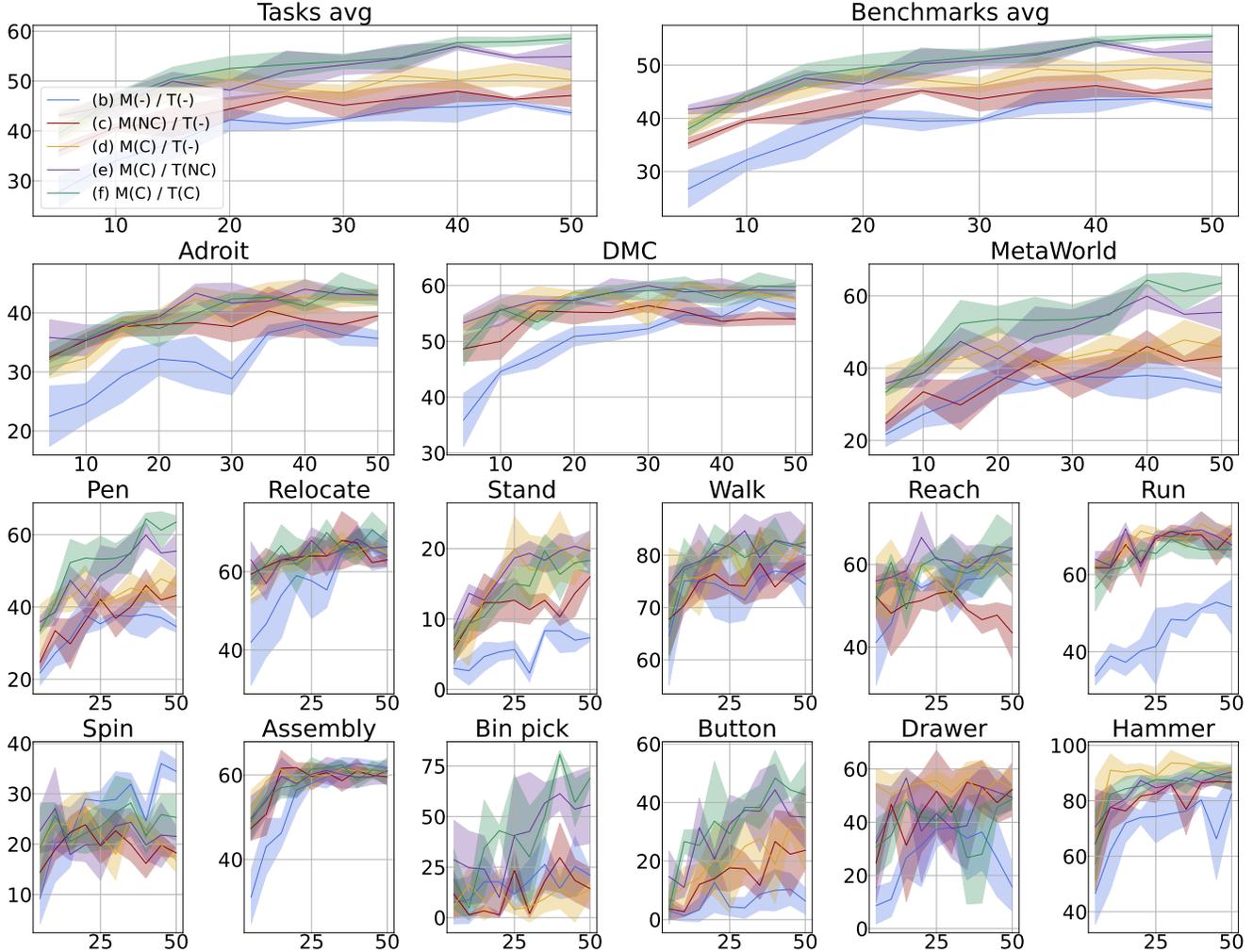}
    \caption{\label{fig:validation_curves} \known~ --- \textbf{Impact of visual adapters}: Evolution of the validation performance during training for rows (b)-(f) in Table 1 of the main paper. In the legend, M and T refer respectively to the state of middle and top adapters, and -, NC or C mean they are absent, not conditioned or conditioned on the task embedding. On all plots, the y-axis represents the performance score and the x-axis corresponds to the training epoch. Colored lines represent the evolution of mean performance over 3 training runs (3 random seeds) and shaded areas represent standard deviation.}
\end{figure*}

Figure~\ref{fig:validation_curves} shows the evolution of the validation score as a function of training epochs for rows (b)-(f) in Table 1 of the main paper. These curves showcase the gain brought by both middle and top adapters, and the positive impact of conditioning them on the task at hand.

\section{Additional ablation studies}
\label{sec:add_ablation_studies}
\myparagraph{Impact of conditioning the policy on the task at hand} Row (b) in Table~\ref{table:additional_ablation_studies} reaches the same performance, even slightly better, as row (a), showing that when adapters are conditioned on the task at hand, conditioning the policy itself is not necessary. This seems to indicate that conditioned adapters already insert task-related information into visual embeddings fed to the policy.

\myparagraph{Using the 'CLS' token representation as input to the policy} Row (c) in Table~\ref{table:additional_ablation_studies} performs worse than row (a), indicating that our introduced tokens aggregation layer $\psi$ improves over the strategy used in previous work consisting in feeding the output 'CLS' token to the policy. This confirms our assumption that the 'CLS' token is undertrained under the MAE pre-training task.

\myparagraph{Impact of the middle adapters when using top adapters} Row (d) in Table~\ref{table:additional_ablation_studies} also performs worse compared with row (a), showing that when training a conditioned top adapter, middle adapters are still very important, bringing a significant boost in performance.

\setlength{\tabcolsep}{1.5pt}
\begin{table*}
    \caption{\label{table:additional_ablation_studies}\known~ --- \textbf{Additional ablation studies}: Validation and test performance on known tasks of different neural variants. Row (a) is equivalent to row (f) in Table 1 of the main paper. When using conditioned adapters, giving the task embedding as input to the policy is not necessary. Our introduced tokens aggregation layer is better than using the representation of the 'CLS' token. Finally, when training a conditioned top adapter, middle adapters are still important and bring a boost in performance. \textbf{Cond $\pi$}: policy conditioned on the task embedding -- C: Conditioned -- \textbf{'CLS' token}: using the 'CLS' token representation as the frame embedding fed to the policy. Performance is reported as \textit{mean $\pm$ std} over $3$ training runs (seeds).}
    \centering
    \small
    {
    \begin{tabular}{l c c c c c c c c c c c c c c c c}
        \toprule
        & \textbf{Cond.} & \multicolumn{2}{c}{\textbf{Adapters}} &  \textbf{'CLS'} & \multicolumn{10}{c}{\textbf{Multi-task performance}} \\
         & \textbf{$\pi$}& Mid. & Top & \textbf{token} & \multicolumn{2}{c}{Adroit} & \multicolumn{2}{c}{DMC} & \multicolumn{2}{c}{MetaWorld} & \multicolumn{2}{c}{Benchmarks avg} & \multicolumn{2}{c}{Tasks avg} \\
         &&&&& Val & Test & Val & Test & Val & Test & Val & Test & Val & Test \\
         \midrule
         (a) & $\checkmark$ & C & C & $-$ & $42.0$ {\scriptsize $\pm$ 0.8} & $42.3$ {\scriptsize $\pm$ 1.0} & $59.9$ {\scriptsize $\pm$ 0.9} & $60.0$ {\scriptsize $\pm$ 0.5} & $65.3$ {\scriptsize $\pm$ 1.0} & $54.5$ {\scriptsize $\pm$ 3.3} & $55.8$ {\scriptsize $\pm$ 0.1} & $52.3$ {\scriptsize $\pm$ 1.0} & $59.2$ {\scriptsize $\pm$ 0.1} & $54.8$ {\scriptsize $\pm$ 1.2} \\
         \midrule
         (b) & $-$ & C & C & $-$ & $42.3$ {\scriptsize $\pm$ 2.0} & $40.8$ {\scriptsize $\pm$ 3.0} & $59.2$ {\scriptsize $\pm$ 1.0} & $59.2$ {\scriptsize $\pm$ 2.3} & $68.7$ {\scriptsize $\pm$ 2.2} & $57.6$ {\scriptsize $\pm$ 3.4} & $56.8$ {\scriptsize $\pm$ 0.8} & $52.5$ {\scriptsize $\pm$ 0.9} & $60.4$ {\scriptsize $\pm$ 0.8} & $55.5$ {\scriptsize $\pm$ 0.5} \\
         \midrule
         (c) & $\checkmark$ & C & C & $\checkmark$ & $38.8$ {\scriptsize $\pm$ 5.9} & $36.2$ {\scriptsize $\pm$ 2.3} & $57.8$ {\scriptsize $\pm$ 1.9} & $58.1$ {\scriptsize $\pm$ 2.6} & $57.5$ {\scriptsize $\pm$ 8.0} & $50.9$ {\scriptsize $\pm$ 4.3} & $51.4$ {\scriptsize $\pm$ 2.8} & $48.4$ {\scriptsize $\pm$ 0.6} & $54.5$ {\scriptsize $\pm$ 2.9} & $51.4$ {\scriptsize $\pm$ 1.3} \\
         \midrule
         (d) & $\checkmark$ & $-$ & C & $-$ & $34.7$ {\scriptsize $\pm$ 1.5} & $34.7$ {\scriptsize $\pm$ 3.3} & $51.4$ {\scriptsize $\pm$ 0.4} & $52.1$ {\scriptsize $\pm$ 0.5} & $53.6$ {\scriptsize $\pm$ 4.4} & $44.5$ {\scriptsize $\pm$ 4.7} & $46.6$ {\scriptsize $\pm$ 1.6} & $43.8$ {\scriptsize $\pm$ 1.8} & $49.5$ {\scriptsize $\pm$ 2.0} & $46.0$ {\scriptsize $\pm$ 1.9} \\
        \bottomrule
    \end{tabular}
    }
\end{table*}

\begin{table*}
    \caption{\label{table:add_vis_backbones}\known~ --- \textbf{Impact on other visual backbones}: Validation and test performance on known tasks for two additional visual backbones (PVR~\cite{parisi2022unsurprising} and MVP~\cite{radosavovic2023real}). Our task-conditioned adapters improve the extracted visual features in both cases, leading to higher multi-task policy performance. Performance is reported as \textit{mean $\pm$ std} over $3$ training runs (seeds).}
    \centering
    \small
    {
    \begin{tabular}{c c c c c c c c c c c c}
        \toprule
         \textbf{ViT} & \textbf{Ours} & \multicolumn{10}{c}{\textbf{Multi-task performance}} \\
         && \multicolumn{2}{c}{Adroit} & \multicolumn{2}{c}{DMC} & \multicolumn{2}{c}{MetaWorld} & \multicolumn{2}{c}{Benchmarks avg} & \multicolumn{2}{c}{Tasks avg} \\
         && Val & Test & Val & Test & Val & Test & Val & Test & Val & Test \\
         \midrule
         \multirow{2}{*}{PVR~\cite{parisi2022unsurprising}} & $-$ & $34.7${\scriptsize ${\pm}$2.8} & $30.2${\scriptsize ${\pm}$ 1.0} & $58.1${\scriptsize ${\pm}$3.0} & $55.3${\scriptsize ${\pm}$ 7.2} & $41.8${\scriptsize ${\pm}$ 1.7} & $33.5${\scriptsize ${\pm}$ 1.4} & $44.8${\scriptsize ${\pm}$ 1.3} & $39.6${\scriptsize ${\pm}$ 3.0} & $47.4${\scriptsize ${\pm}$ 1.3} & $42.0${\scriptsize ${\pm}$ 3.5} \\
         
          & $\checkmark$ & $\textbf{43.3}${\scriptsize ${\pm}$ 2.5} & $\textbf{41.0}${\scriptsize ${\pm}$ 5.1} & $\textbf{61.9}${\scriptsize ${\pm}$ 1.3} & $\textbf{61.3}${\scriptsize ${\pm}$ 1.2} & $\textbf{66.8}${\scriptsize ${\pm}$ 2.3} & $\textbf{55.7}${\scriptsize ${\pm}$ 3.4} & $\textbf{57.3}${\scriptsize ${\pm}$ 1.0} & $\textbf{52.7}${\scriptsize ${\pm}$ 3.1} & $\textbf{60.9}${\scriptsize ${\pm}$ 0.8} & $\textbf{55.6}${\scriptsize ${\pm}$ 2.7} \\
        \midrule
        \multirow{2}{*}{MVP~\cite{radosavovic2023real}} & $-$ & $38.0${\scriptsize ${\pm}$1.3} & $34.3${\scriptsize ${\pm}$2.4} & $56.5${\scriptsize ${\pm}$2.1} & $56.5${\scriptsize ${\pm}$1.7} & $42.8${\scriptsize ${\pm}$6.9} & $35.9${\scriptsize ${\pm}$5.6} & $45.8${\scriptsize ${\pm}$1.5} & $42.2${\scriptsize ${\pm}$2.2} & $47.7${\scriptsize ${\pm}$1.9} & $44.2${\scriptsize ${\pm}$2.2} \\
        
         & $\checkmark$ & $\textbf{47.7}${\scriptsize ${\pm}$5.9} & $\textbf{46.2}${\scriptsize ${\pm}$2.4} & $57.3${\scriptsize ${\pm}$2.9} & $56.9${\scriptsize ${\pm}$2.5} & $\textbf{64.9}${\scriptsize ${\pm}$12.1} & $\textbf{55.4}${\scriptsize ${\pm}$12.2} & $\textbf{56.6}${\scriptsize ${\pm}$4.0} & $\textbf{52.8}${\scriptsize ${\pm}$2.6} & $\textbf{58.9}${\scriptsize ${\pm}$4.4} & $\textbf{54.5}${\scriptsize ${\pm}$3.8} \\
        \bottomrule
    \end{tabular}
    }
\end{table*}

\begin{table}[t]
    \caption{\label{table:actions_probing}\textbf{Non-linear probing of actions}: we explore the performance of action regression from the visual embedding of a single frame. Considered metrics are the Mean Squared Error (MSE) and coefficient of determination ($R^2$). The top adapter seems to insert action-related information into the visual embedding, as the probing MLP achieves the best performance.}
    \centering
    \begin{tabular}{l c c c c}
        \toprule
          & Middle adapters & Top adapter & MSE & $R^2$ \\
         \midrule
         (a) & $-$ & $-$ & $0.067$ & $0.69$ \\
         (b) & NC & $-$ & $0.069$ & $0.57$ \\
         (c) & C & $-$ & $0.069$ & $0.59$ \\
         (d) & C & NC & $0.037$ & $0.90$ \\
         (e) & C & C & $\textbf{0.034}$ & $\textbf{0.92}$ \\
        \bottomrule
    \end{tabular}
\end{table}

\begin{table}
    \caption{\label{table:MetaWorld_only}\known~ --- \textbf{Diversity of known tasks}: Validation and test performance on MetaWorld known tasks when our approach with task-conditioned adapters is either trained on the three considered benchmarks (Adroit, DMC, MetaWorld), or on tasks from MetaWorld only. As expected, the model trained on known tasks from MetaWorld only reaches higher performance. Performance is reported as \textit{mean $\pm$ std} over $3$ training runs (seeds).}
    \centering
    \begin{tabular}{l c c c}
        \toprule
        & \textbf{Training} & \multicolumn{2}{c}{\textbf{MetaWorld}} \\
        && Val & Test \\
        \midrule
         (a) & All 3 benchmarks  & $65.3${\scriptsize{$\pm$}1.0} & $54.5$ {\scriptsize{$\pm$}3.3} \\
         (b) & MetaWorld only & $75.6${\scriptsize{$\pm$}1.6} & $67.8${\scriptsize{$\pm$}2.6} \\
        \bottomrule
    \end{tabular}
\end{table}

\begin{table*}
    \caption{\label{table:fewshot_finetuned_baseline}\unknown~ --- Performance of a finetuned baseline (\textit{Ft.}) and task embedding search (\textit{TE opt.}) for a policy either trained on MetaWorld only (\textit{MV}) or all 3 benchmarks (\textit{All 3}). $t_i^u$ refers to the $i$-th unknown task. Performance is reported as \textit{mean $\pm$ std} over $3$ training runs (seeds).}
    \centering
    \small
    {
    \begin{tabular}{c c c c c c c c c c c c c c c c c c c c}
        \toprule
        & \textbf{Opt.} & \textbf{Train.} & \textbf{Setting} & $t_0^u$ & $t_1^u$ & $t_2^u$ & $t_3^u$ & $t_4^u$ & $t_5^u$ & $t_6^u$ & $t_7^u$ & $t_8^u$ & $t_9^u$ & $t_{10}^u$ & $t_{11}^u$ & $t_{12}^u$ & $t_{13}^u$ & $t_{14}^u$ & \textbf{Mean}  \\
        \midrule
         (a) & TE opt. & All 3 & Single policy & $2${\scriptsize ${\pm}$4} & $1${\scriptsize ${\pm}$2} & $55$ {\scriptsize ${\pm}$26} & $41${\scriptsize ${\pm}$21} & $4${\scriptsize ${\pm}$4} & $34${\scriptsize ${\pm}$10} & $81${\scriptsize ${\pm}$21} & $72${\scriptsize ${\pm}$12} & $0${\scriptsize ${\pm}$0} & $11${\scriptsize ${\pm}$17} & $34${\scriptsize ${\pm}$30} & $48${\scriptsize ${\pm}$6} & $47${\scriptsize ${\pm}$2} & $53${\scriptsize ${\pm}$21} & $4${\scriptsize ${\pm}$1} & $33${\scriptsize ${\pm}$2} \\
         (b) & Ft. & All 3 & $15$ policies & $1${\scriptsize ${\pm}$2} & $0${\scriptsize ${\pm}$0} & $49${\scriptsize ${\pm}$44} & $69${\scriptsize ${\pm}$22} & $5${\scriptsize ${\pm}$2} & $62${\scriptsize ${\pm}$8} & $96${\scriptsize ${\pm}$4} & $91${\scriptsize ${\pm}$7} & $2${\scriptsize ${\pm}$3} & $54${\scriptsize ${\pm}$8} & $50${\scriptsize ${\pm}$4} & $22${\scriptsize ${\pm}$19} & $66${\scriptsize ${\pm}$7} & $89${\scriptsize ${\pm}$1} & $3${\scriptsize ${\pm}$2} & $44${\scriptsize ${\pm}$3} \\
         \midrule
         (c) & TE opt. & MW & Single policy & $3${\scriptsize ${\pm}$6} & $0${\scriptsize ${\pm}$0} & $56${\scriptsize ${\pm}$44} & $64${\scriptsize ${\pm}$20} & $1${\scriptsize ${\pm}$2} & $69${\scriptsize ${\pm}$19} & $100${\scriptsize ${\pm}$0} & $77${\scriptsize ${\pm}$20} & $0${\scriptsize ${\pm}$1} & $6${\scriptsize ${\pm}$6} & $22${\scriptsize ${\pm}$38} & $19${\scriptsize ${\pm}$3} & $55${\scriptsize ${\pm}$13} & $57${\scriptsize ${\pm}$17} & $5${\scriptsize ${\pm}$3} & $36${\scriptsize ${\pm}$5} \\
        \bottomrule
    \end{tabular}
    }
\end{table*}

\section{Impact on other visual backbones} 
\label{sec:add_visual_backbones}
Table~\ref{table:add_vis_backbones} shows the impact of our task-conditioned adapters on the visual features extracted by two other SOTA ViT-B backbones, i.e. PVR~\cite{parisi2022unsurprising} and MVP~\cite{radosavovic2023real}. As can be seen, our adapters bring a boost in policy performance for both pre-trained backbones, generalizing the conclusions obtained with VC-1.

\section{Visualizing the influence of task-conditioned adaptation}
This section focuses on investigating the impact of the introduced adapters on the processing of visual features. All sequences of visual frames used in the following experiments are taken from a held-out set of expert trajectories not used at training time.

\begin{figure*}[t]
    \centering
    \includegraphics[width=\textwidth]{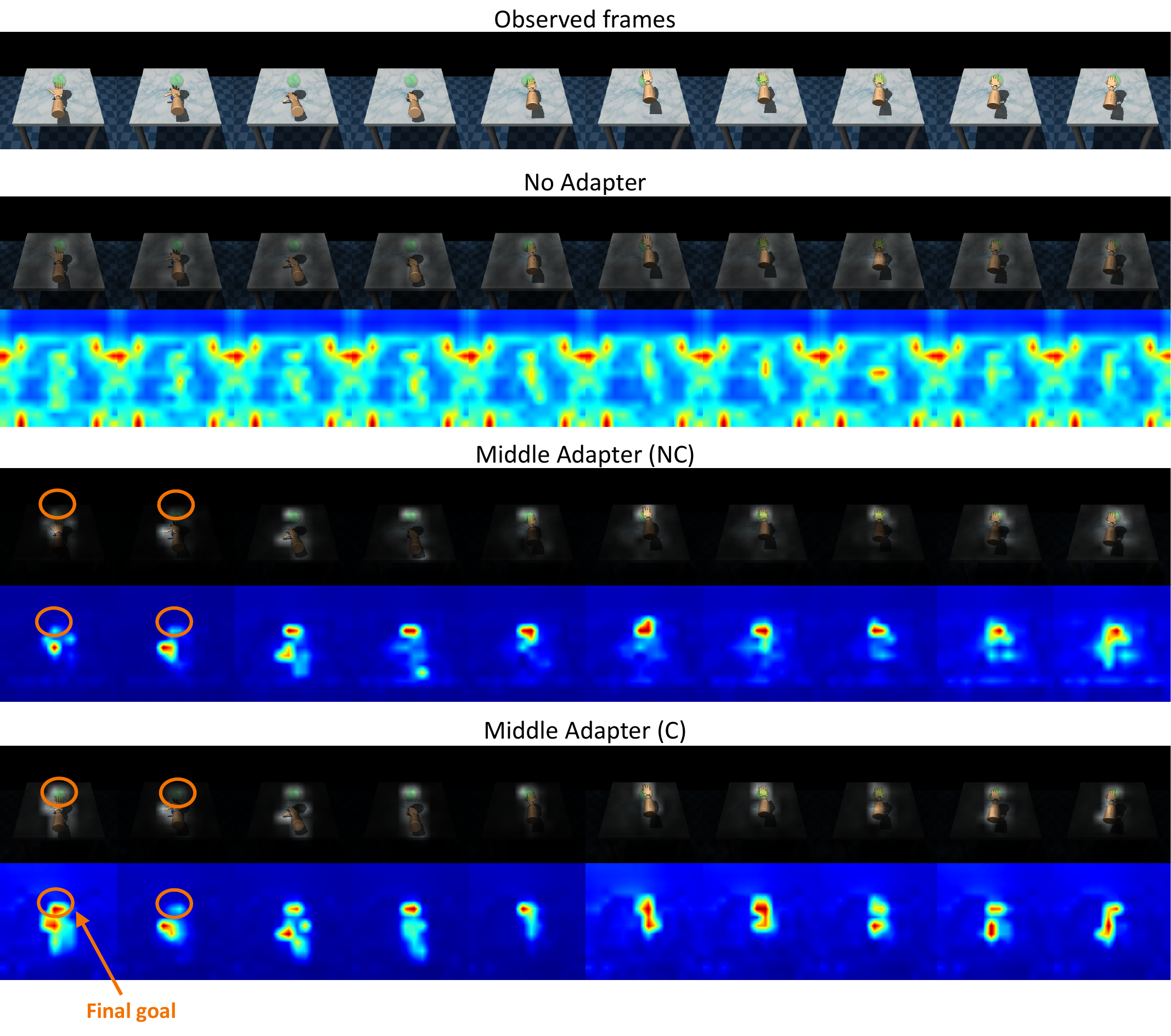}
    \caption{\label{fig:viz_attention_relocate} \textbf{Visualization of attention maps (Relocate task)}. First row: observed input frames. Following blocks: for each model type, we show the attention map of the last ViT layer, first overlaid on top of the visual frame and below as a colored heatmap. In this example, middle adapters allow to focus the attention on important regions, and task conditioning leads to a better covering of the robotic hand and the sphere goal in all frames.}
\end{figure*}

\myparagraph{Influence of middle adapters on ViT attention maps} Figure~\ref{fig:viz_attention_relocate} presents another visualization (see also Figure~\ref{fig:viz_attention_assembly} in the main paper) of the attention map of the last layer of the vision encoder: we sum attention maps for all tokens and all heads, and normalize them between $0$ and $1$. The first row shows a sequence of visual frames and below, for each model variant (No Adapter, Middle Adapter (NC), Middle Adapter (C)), one can see the attention map overlaid on top of the visual frame and displayed below as a colored heatmap.

Figure~\ref{fig:viz_attention_relocate} confirms that middle adapters lead to better-focused attention around important objects related to the task, compared with vanilla VC-1. Unconditioned adapters (NC) tend to either produce very narrow (first frames in Figure~\ref{fig:viz_attention_relocate}) or quite broad (last frames in Figure~\ref{fig:viz_attention_relocate}) attention. Task conditioning leads to a better coverage of entire objects and important agent parts. As already mentioned in the main paper, conditioning on the task at hand improves the attention towards the final goal to reach.

\begin{figure*}[t]
    \centering
    \begin{minipage}{.6\textwidth}
        \includegraphics[width=\textwidth]{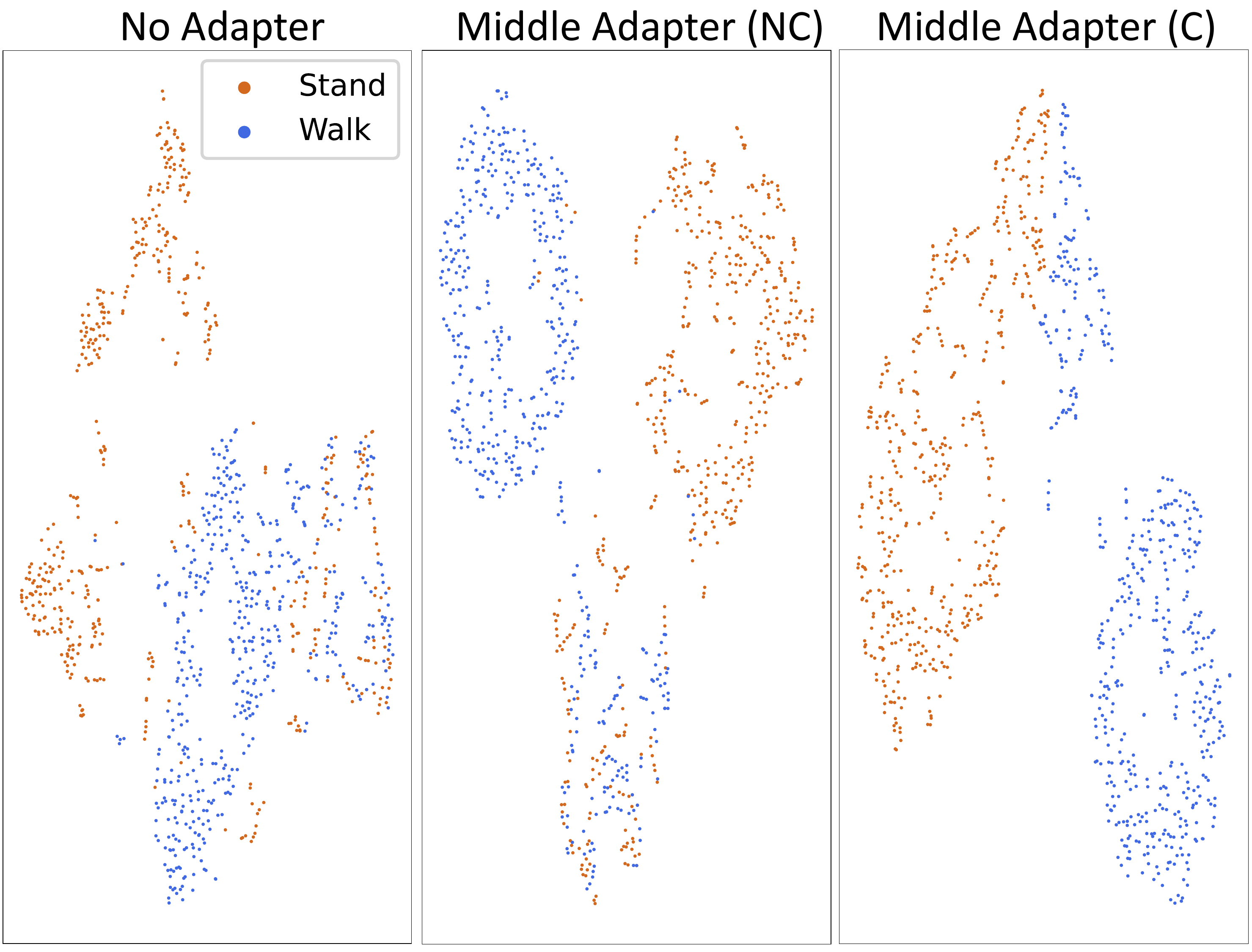}
        \caption{\label{fig:tsne_stand_walk} \textbf{Task-related information inside visual embeddings}: t-SNE plots of visual embeddings for a set of frames for DMC \textit{Stand} and \textit{Walk} tasks. We chose these tasks for their visual similarity, making it very hard to distinguish between them from vision only. Conditioning of the middle adapters leads to two properly separated clusters, showing the insertion of task-related information into the visual embeddings.}
    \end{minipage}
    \hfill
    \begin{minipage}{.35\textwidth}
        \includegraphics[width=\textwidth]{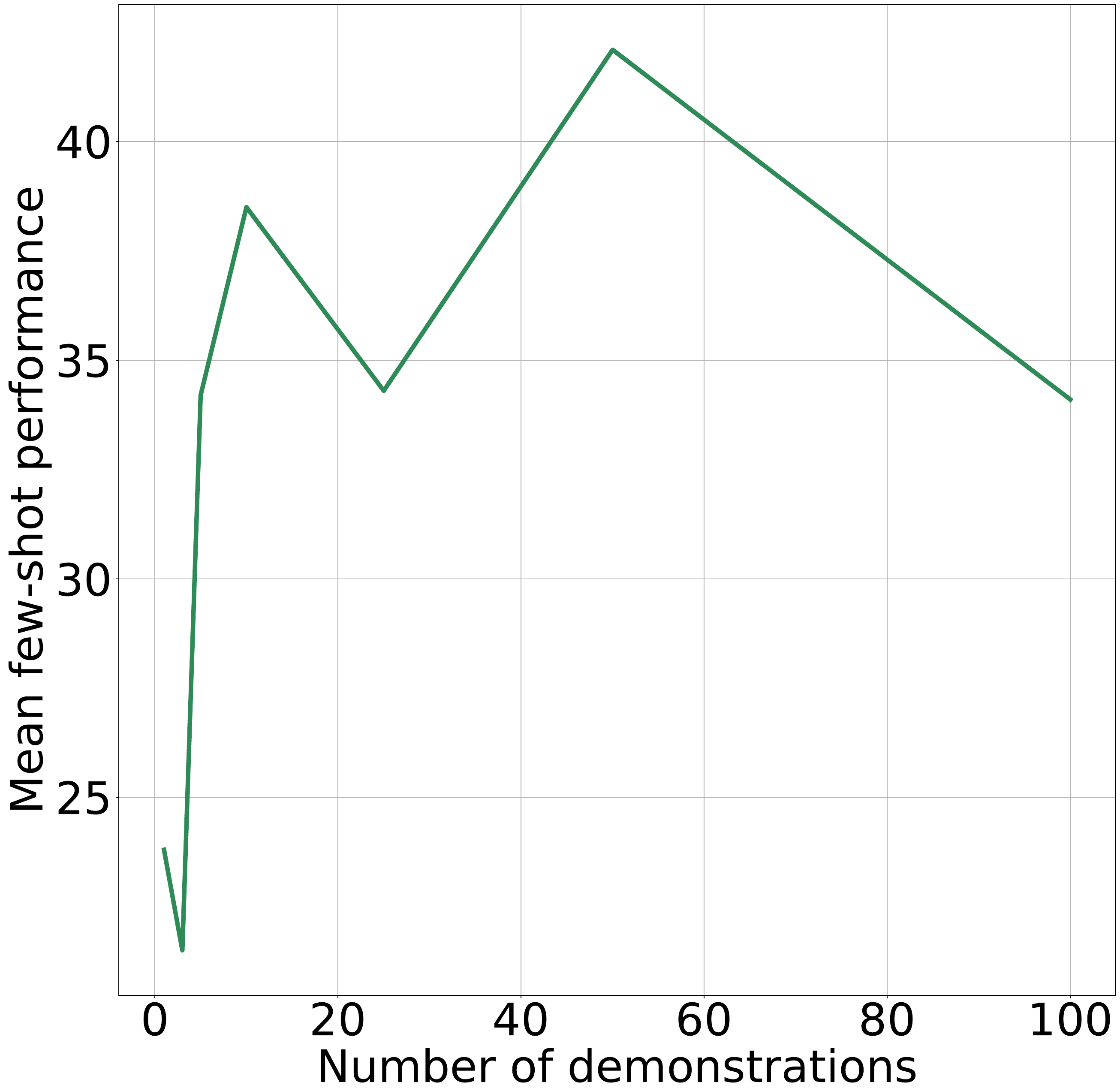}
        \caption{\label{fig:impact_nb_demos_few_shot_perf} \unknown~ --- \textbf{Impact of the number of demonstrations on few-shot adaptation to unseen tasks}: Few-shot performance as a function of the number of demonstrations used to optimize the task embedding. We can achieve $23.8\%$ from a single demonstration, and more demonstrations can lead to higher performance. However, the trend is not as simple as $100$ demonstrations lead to the same performance as $5$ demonstrations.}
    \end{minipage}
\end{figure*}

\myparagraph{Conditioning middle adapters helps insert task-related information into visual embeddings} In order to study the underlying mechanisms of adapter modules, we examine the content of produced visual embeddings. Figure~\ref{fig:tsne_stand_walk} shows t-SNE plots of visual embeddings for a set of frames for both the DMC \textit{Stand} and \textit{Walk} tasks. Visual observations are identical for both tasks at the beginning of rollouts, and very similar in the rest of the sequences, making it very hard to distinguish between these $2$ tasks from a visual observation only. Embeddings from the conditioned middle adapters form two well-separated clusters, showcasing the task-related information brought by conditioning adapters on the task at hand.

\section{Non-linear probing of actions} Table~\ref{table:actions_probing} shows the performance of a probing MLP network trained to regress the expert action to take from the visual embedding of a single frame only. As can be seen, its performance improves drastically when trained on embeddings predicted by a vision encoder composed of conditioned middle and top adapters. A conditioned top adapter thus inserts action-related information within visual embeddings.

\section{Diversity of known tasks} Table~\ref{table:fewshot_finetuned_baseline} (c) and Table~\ref{table:MetaWorld_only} (b) show a model trained on MetaWorld only, which performs better on MetaWorld than models trained on all 3 benchmarks (the domain gap between them is large). The lower performance on MetaWorld when training on all 3 benchmarks is largely outweighed by the ability to address Adroit and DMC.

\section{Few-shot adaptation baseline} Table~\ref{table:fewshot_finetuned_baseline} compares model finetuning on new tasks (b) with our task embedding search (a). As expected, (b) performs better but task embedding search (a) solves a harder problem, as we keep a single policy. Our adapters can thus be used in $2$ settings: (i) task embedding search, keeping a single policy addressing all tasks (low memory footprint), (ii) task-specific fine-tuning to reach the best performance possible if memory is not an issue (one specific set of $130$M parameters for each task). 

\section{Architecture details}
\myparagraph{Vision encoder} we use a ViT-B backbone, initialized from VC-1 weights, as the base vision encoder $\phi$. It is made of $12$ self-attention layers, each composed of $12$ attention heads, with a hidden size of $768$. The input image of size $224{\times}224{\times}3$ is divided into a grid of $14{\times}14$ patches, where each patch has thus a size of $16{\times}16$ pixels. An additional 'CLS' token is appended to the sequence of image tokens to follow the setup used to pre-train the model.

\myparagraph{Task embedding} the task embedding is a $1024$-dim vector. For \textit{known tasks}, it is predicted by a linear embedding layer from a 1-in-K vector where $K{=}12$.

\myparagraph{Middle adapters} one adaptation module $\alpha_l$ is inserted after each self-attention layer inside $\phi$. It is composed of $2$ fully-connected layers with respectively $384$ and $768$ neurons. A \textit{GELU} activation function is applied to the output of the first layer. The input to a middle adapter is the concatenation of the task embedding and a token representation from the previous self-attention layer. It thus processes all tokens as a batch.

\myparagraph{Aggregation fully-connected layer} the input to the aggregation fully-connected layer $\psi$ is a concatenation of the $768$-dim representation of all $14{\times}14{=}196$ tokens. It is implemented as a simple fully-connected layer predicting a $768$-dim vector representation.

\myparagraph{Top adapter} the top adapter $\tau$ is fed with the output of $\psi$, again concatenated to the task embedding. It is composed of $2$ fully-connected layers that both have $768$ neurons. A \textit{ReLU} activation function is applied to the output of the first layer.

\myparagraph{Multi-task policy} The policy $\pi^m$ is a $3$-layer MLP, with $256$ neurons for all layers and $ReLU$ activation functions. A batch normalization operation is applied to the input to the policy. $\pi^m$ outputs a $30$-dim action vector, as $30$ is the number of components in the action space with the most components among the $12$ known tasks. When solving a task with a smaller action space, we mask out the additional dimensions.

\section{Impact of the number of demonstrations on few-shot adaptation to unseen tasks}
Figure~\ref{fig:impact_nb_demos_few_shot_perf} presents the evolution of the average few-shot performance of our method across the $15$ unknown tasks depending on the number of available demonstrations when optimizing the task embedding. From only a single demonstration per task, we can already reach a satisfying $23.8\%$ mean performance. Adding more demonstrations can allow to reach higher performance, but the scaling law does not appear to be as simple as using $100$ demonstrations leads to the same final performance as $5$ demonstrations.

\end{document}